\pdfoutput=1
\documentclass[10pt,twocolumn,letterpaper]{article}

\usepackage{wacv}

\usepackage{graphicx}
\graphicspath{{figures/}}
\usepackage{amsmath,amssymb,amsthm}
\usepackage{booktabs}
\usepackage{multirow}
\usepackage{xcolor}
\usepackage{xspace}
\usepackage[accsupp]{axessibility}

\definecolor{wacvblue}{rgb}{0.21,0.49,0.74}
\usepackage[pagebackref,breaklinks,colorlinks,allcolors=wacvblue]{hyperref}
\hypersetup{pdftitle={Discarded or Unreachable? Diagnosing and Repairing Spatial Relation Failure in Vision-Language Similarity}, pdfauthor={Mahdi Mostajabdaveh, Zirui Zhou}}

\def\wacvPaperID{3085}
\def\confName{WACV}
\def\confYear{2027}

\newtheorem{proposition}{Proposition}
\newtheorem{lemma}{Lemma}

\newcommand{\zm}{z^{-}}
\newcommand{\zp}{z^{+}}
\newcommand{\moment}{spatial moment\xspace}

\begin{document}

\title{Discarded or Unreachable? Diagnosing and Repairing\\
Spatial Relation Failure in Vision--Language Similarity}

\author{Mahdi Mostajabdaveh\\
Huawei Technologies Canada
\and
Zirui Zhou\\
Huawei Technologies Canada
}
\maketitle

\begin{abstract}
Contrastive vision--language models score left/right relations near chance, but the symptom has two
causes that call for opposite repairs: the relation may survive in the pooled embedding yet be
unreachable by native image--text similarity, or pooling may discard it while the dense features
retain it. On clean two-object photographs, across ten encoders, the lateral polarity is linearly
decodable from the flip-odd component of the pooled vector at $0.77$--$0.99$, yet a search over a
million captions leaves nine of the ten at chance under native scoring: an access failure. On
controlled $C_4$ orbits, where the relation is the only thing that varies, pooling keeps the
arrangement axis at $0.998$--$1.000$ and loses its polarity, which the patch grid retains: a pooling
failure. Both sit downstream of the dense features, so one readout covers both. Our zero-parameter,
query-conditioned spatial moment recomputes the geometry from the patch grid and stores it as a
vector beside the native embedding. On a $118{,}287$-image index it raises left/right pairwise
accuracy against reversed-geometry negatives from $0.51$--$0.53$ to $0.81$--$0.96$ and conditional
R@1 among $21$ category-matched candidates from $0.05$ to $0.37$--$0.75$, at under a millisecond
per batched query, while reproducing the native ranking exactly for queries that name no
relation.\footnote{Code, stimulus generators and all result files:
\url{https://github.com/mahdims/vl-spatial-moment}.}
\end{abstract}

\section{Introduction}
\label{sec:intro}

\begin{figure*}[t]
\centering
\includegraphics[width=\textwidth]{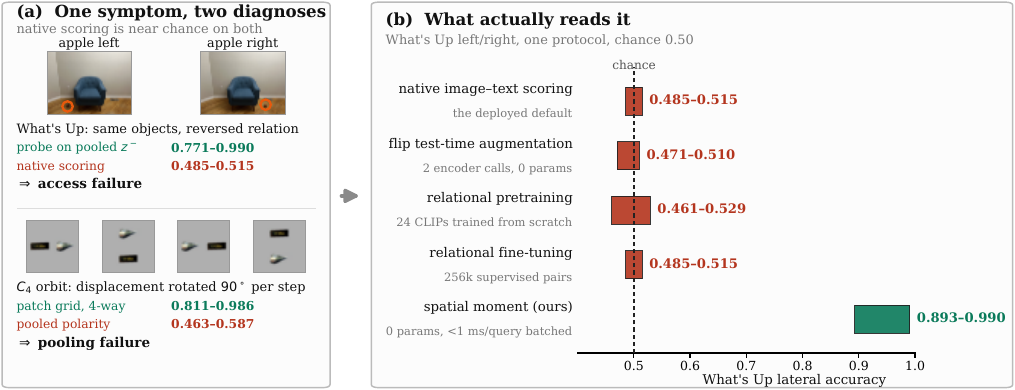}
\caption{\textbf{One near-chance symptom, two failures, one repair.} (a) On What's~Up the pooled
vector carries the polarity, in its flip-odd component $\zm$ (Sec.~\ref{sec:framework}), but native
scoring cannot reach it; on the $C_4$ orbits only the patch grid carries it. Chance is $0.5$ ($0.25$
for the 4-way row); the ring marks the object that moves. (b) What's~Up left/right under one
protocol. Flip test-time augmentation compares a photograph with its mirror; pretraining is $24$
ViT-B/32 CLIPs trained from scratch with up to all captions relational; fine-tuning is three
fine-tunes on $256$k pairs. Ranges are over the ten encoders of Tab.~\ref{tab:main}, or over runs.}
\label{fig:story}
\end{figure*}

Ask an image search engine for \emph{a dog to the left of a sofa} and it returns dogs and sofas in
either arrangement: the relation contributes nothing to the ranking. Contrastive dual encoders such
as CLIP~\cite{clip} and SigLIP~2~\cite{siglip2}, which embed image and caption separately, score them
by an inner product and underlie vector-index retrieval, identify \emph{what} is in an image well and
\emph{where} things sit relative to each other poorly, left/right worst~\cite{whatsup,vsr,aro};
more data, larger models and patch--text pretraining do not close the gap. They are this paper's
subject.

Chance accuracy is ambiguous. It is consistent with an encoder that never represented the relation
and with one that represents it somewhere the similarity function cannot reach, and the two call for
opposite repairs: the first needs new information, the second a different reader. We separate them
(Fig.~\ref{fig:story}) for the \emph{polarity} of a lateral relation: which of two objects is on the
left. An encoder computes a grid of patch features and \emph{pools} them,
through a class token, an attention head or an average, into the one vector that retrieval indexes
and compares. In an \emph{access failure} the polarity survives pooling, sits in a component we can
name exactly, and is unreadable by the inner product the system runs. In a \emph{pooling failure}
the pooled vector does not retain the polarity, but the patch features it was pooled from still do.

Mirroring an image exchanges left and right, so we split each pooled embedding into the part the
mirror leaves unchanged and the part it negates, its \emph{flip-odd} component
(Sec.~\ref{sec:framework}). Every linear reader that tells an image from its own mirror reads only
the flip-odd component (Lemma~\ref{lem:odd}), and so does flip test-time augmentation
(Prop.~\ref{prop:tta}). ``Is the polarity in the pooled vector?'' then splits into two questions with
separate answers: is it in the flip-odd component, and can a caption reach it?

On clean photographs the answers are yes and no. A linear probe reads the lateral relation from the
flip-odd component at $0.771$--$0.990$, while native scoring is at chance on all ten encoders under
ranking protocols where a fixed caption preference cannot help. A bad prompt is the obvious
explanation, so we search for a good one. The best bias-free direction in the caption subspace
reaches $0.69$--$0.97$ and real relational phrasings $0.48$--$0.56$, near-orthogonal to the
image-side direction on every encoder; a search two orders of magnitude larger leaves nine encoders
at chance and finds a caption on the tenth (Sec.~\ref{sec:bound}).

On controlled stimuli the polarity does not survive pooling. A $C_4$ orbit pastes two objects on a
plain background and rotates the displacement between them by $90^\circ$ per step, so its lateral
pair holds the same two positions and differs only in which object sits where. The pooled vector identifies the
arrangement axis at $0.998$--$1.000$ and reads the polarity along it near chance on nine of the ten
encoders, under every linear and nonlinear reader we trained, while the patch grid of the same
images reads the full 4-way relation at $0.811$--$0.986$ (Sec.~\ref{sec:c4}).

\vspace{2pt}
\noindent\textbf{Contributions.}
\begin{itemize}\itemsep2pt
  \item \textbf{Tests that separate access failure from pooling failure.} Flip symmetry and
  caption-space geometry expose decodable-but-unreached laterality on real photographs, exact $C_4$
  orbits show pooling drops the polarity, and a closed-form model predicts native accuracy from the
  measured geometry (Secs.~\ref{sec:framework}--\ref{sec:c4}).

  \item \textbf{A training-free repair derived from that diagnosis.} Both failures sit downstream of
  the patch grid, so a zero-parameter spatial moment, read through each encoder's own pooling head,
  recomputes the geometry from it and stores it as a vector beside the native embedding; the
  repaired score stays one inner product served from an ordinary vector index. On a
  $118{,}287$-image index it raises left/right pairwise accuracy against reversed-geometry negatives
  from $0.51$--$0.53$ to $0.81$--$0.96$, and leaves spatially empty rankings unchanged
  (Secs.~\ref{sec:bridge}--\ref{sec:retrieval}). Relational supervision, the standard alternative,
  does not restore native reachability in the forms and at the scales we tested
  (Sec.~\ref{sec:supervision}).
\end{itemize}

\section{Related Work}
\label{sec:related}

\noindent\textbf{The deficit, measured at the output.} What's~Up~\cite{whatsup}, VSR~\cite{vsr} and
ARO~\cite{aro} establish that VL models handle spatial relations poorly and that left/right is the
worst case, and COMFORT~\cite{comfort} adds frame-of-reference ambiguity. Compositional benchmarks
are often solvable blind~\cite{crepeaudit}, which our controls address (Sec.~\ref{sec:measure}). All
of it reads the emitted score rather than the vector producing it, so none can say whether the
relation is absent or out of reach.

\noindent\textbf{Decodable versus used.} That a probe recovers a property does not mean the model
uses it; amnesic probing~\cite{amnesic}, closed-form erasure~\cite{leace} and
$\mathcal{V}$-information~\cite{vinformation} were built for that gap. We instantiate the reader
family as the inner products deployed retrieval actually computes, which gives ``unused'' a bound
instead of an intuition. A spatial relation is a case of the binding problem~\cite{greffbinding}.
Contrastive embeddings compose concepts nearly additively, and binding generalises only as far as
training covers the combinations~\cite{uselisscaling,uselisbind}; no CLIP-like joint space can
represent spatial relations together with any one of content, attribute binding or
negation~\cite{dcsm}. The same encoded-but-unaligned shape is reported for attribute
binding~\cite{bagofwords_crossmodal,uselisbind} and negation~\cite{negationwhere,negationlens}, and
contrastive training provably admits encoders that align well while staying insensitive to
compositional edits~\cite{tokencausal}; none of it measures the geometry between the towers for a
spatial relation.

\noindent\textbf{Dense features, symmetry, and retraining.} Both ingredients of the repair already
exist. Group averaging is standard for equivariant design~\cite{frameaveraging};
MirrorCLIP~\cite{mirrorclip} runs the original-versus-flipped comparison and \emph{discards} that
channel as a rendered-text artifact. Coordinate-weighted feature averages are
old~\cite{spatialautoencoder}, frozen features retain locality well enough for
detection~\cite{fvlm}, and lightweight modules~\cite{refiningclip}, SigLIP~2~\cite{siglip2}, and
training-free changes to the last attention block~\cite{maskclip,sclip,clearclip} recover dense
structure. DCSM~\cite{dcsm} trains a scoring head on each image--caption pair's dense similarity map,
which needs both grids at query time. A further line argues spatial competence requires changed
pretraining~\cite{lostinspace,spatialclip,tips}. Retraining is the other standard
answer~\cite{aro,whatsleft,lrsymmetry}, and all of it takes supervision to be the missing
ingredient. None of it asks which failure mode dense features remedy, whether the moment reads a
different signal from the pooled one (it does; Appendix~\ref*{app:bridge}), or whether supervision
was ever the constraint (Sec.~\ref{sec:supervision}).

\section{Reachability Under Native Scoring}
\label{sec:framework}

Let $z(I) \in \mathbb{R}^D$ be the deployed, $L_2$-normalised image embedding and $t(c)$ the text
embedding of caption $c$. Native scoring over a caption pair decides by
$\mathrm{sign}\langle z(I), t(c_1) - t(c_2)\rangle$; retrieval ranks images for a fixed $c$ by
$\langle z(I), t(c)\rangle$. Both are inner products against a \emph{realizable} caption embedding,
$t(c)$ for an actual string $c$ rather than any vector in the span of such embeddings, and that
restriction is what this paper studies.

\vspace{2pt}
$\mathcal{V}$-information~\cite{vinformation}
formalises information relative to a family of readers. With $\mathcal{V}_{\text{lin}}$ all linear
probes and $\mathcal{V}_{\text{nat}} = \{z \mapsto \mathrm{sign}\langle z, t(c_1)-t(c_2)\rangle\}$
over realizable caption pairs, $\mathcal{V}_{\text{nat}} \subset \mathcal{V}_{\text{lin}}$ gives
$I_{\mathcal{V}_{\text{nat}}} \le I_{\mathcal{V}_{\text{lin}}}$. That gap, a property present and a
metric unable to use it, is what we call an \emph{access failure}: the polarity is linearly
decodable from $z$ and unreached by the caption family we search. Reachability is estimated from a
finite set of realizable captions, so the claim is about the family searched, never about absence.

\noindent\textbf{Symmetry channels of the horizontal flip.} $F$ denotes horizontal flip, $F^2 = \mathrm{id}$. The projections onto the trivial and sign
representations, the isotypic decomposition~\cite{frameaveraging} applied to a frozen encoder's
output, are
\[
  \zp(I) = \tfrac{1}{2}\big(z(I) + z(FI)\big), \qquad
  \zm(I) = \tfrac{1}{2}\big(z(I) - z(FI)\big),
\]
with $\zm(FI) = -\zm(I)$.

\begin{lemma}[Mirror discrimination lives in $\zm$]
\label{lem:odd}
For any $w$, $\langle w, z(I)\rangle - \langle w, z(FI)\rangle = 2\langle w, \zm(I)\rangle$.
Consequently every linear reader that separates an image from its own mirror does so through $\zm$,
and $\zp$ contributes exactly zero.
\end{lemma}

The proof is immediate. On mirror pairs $\zp(I) = \zp(FI)$ identically, so any classifier gives
both members of a pair the same label and is right on exactly one of them: a probe on $\zp$ scores
exactly $0.500$ on every fold, not only on average. We check it numerically, where it doubles as a
correctness test on flip handling, batching and normalisation.

\begin{proposition}[Flip TTA is the flip-odd score]
\label{prop:tta}
$\langle z(I), t\rangle > \langle z(FI), t\rangle \iff \langle \zm(I), t\rangle > 0$.
\end{proposition}

So ranking an image above its own mirror \emph{is} the flip-odd score; it needs two embedding
calls, works through a black-box API, and is verified item by item.

\noindent\textbf{Directions realizable by captions.}
\label{sec:bounddef}
Let $V = \mathrm{span}\{t(c) : c \text{ realizable}\}$, estimated from a large caption corpus, and
$V_k$ its top-$k$ principal subspace. In the paired protocol the true image outranks the false one
iff $\langle z(I_{\text{t}}) - z(I_{\text{f}}), t(c)\rangle > 0$, a \emph{single direction with no
intercept}. The matched bound is therefore the best accuracy any unit direction in $V_k$ achieves on
those difference vectors. Bias-free directions are what make it a measure of alignment rather than
of supervision.

\noindent\textbf{A decomposition of the native margin.} With $\Delta\mu = \mathbb{E}[z \mid \text{left}] - \mathbb{E}[z \mid \text{right}]$ and
$\delta_T = t(c_1) - t(c_2)$, the native mean margin factorises as
\begin{equation}
  \langle \Delta\mu, \delta_T\rangle
  = \underbrace{\|\Delta\mu\|}_{\text{encoding}} \cdot
    \underbrace{\|\delta_T\|}_{\text{text contrast}} \cdot
    \underbrace{\cos\angle(\Delta\mu, \delta_T)}_{\text{alignment}} .
  \label{eq:decomp}
\end{equation}
Failure can sit in any factor, and each calls for a different repair;
Sec.~\ref{sec:bound} shows the lateral case is alignment.

\section{Access Failure on Real Photographs}
\label{sec:measure}

\begin{table*}[t]
\centering\footnotesize
\setlength{\tabcolsep}{4pt}
\begin{tabular}{@{}llcccccccc@{}}
\toprule
& & \multicolumn{4}{c}{Supervised probe (What's~Up, lateral)} & \multicolumn{2}{c}{Mirror pairs} & \multicolumn{2}{c}{Mirror ranking} \\
\cmidrule(lr){3-6}\cmidrule(lr){7-8}\cmidrule(lr){9-10}
Encoder & Pooling & $z$ & $\zm$ & $\Delta_{\text{lat}}$ & $\Delta_{\text{vert}}$ & $\zm$ & $\zp$ & flip effect & moment \\
\midrule
CLIP ViT-B/16 & CLS + proj & 0.674 & \textbf{0.771} & $+0.097$ & $-0.015$ & 1.000 & 0.500 & 0.510 [0.476, 0.549] & \textbf{0.951} \\
CLIP LAION-2B & CLS + proj & 0.705 & \textbf{0.878} & $+0.174$ & $-0.005$ & 1.000 & 0.500 & 0.500 [0.461, 0.539] & \textbf{0.942} \\
CLIP DataComp & CLS + proj & 0.713 & \textbf{0.805} & $+0.092$ & $-0.015$ & 1.000 & 0.500 & 0.495 [0.466, 0.524] & \textbf{0.893} \\
MetaCLIP-B & CLS + proj & 0.700 & \textbf{0.815} & $+0.115$ & $+0.000$ & 0.990 & 0.500 & 0.510 [0.485, 0.534] & \textbf{0.951} \\
DFN-B & CLS + proj & 0.660 & \textbf{0.825} & $+0.165$ & $+0.005$ & 1.000 & 0.500 & 0.510 [0.490, 0.534] & \textbf{0.942} \\
SigLIP2-B & attn-pool (MAP head) & 0.698 & \textbf{0.863} & $+0.165$ & $-0.005$ & 1.000 & 0.500 & 0.490 [0.471, 0.510] & \textbf{0.990} \\
SigLIP2-L & attn-pool (MAP head) & 0.660 & \textbf{0.830} & $+0.170$ & $-0.010$ & 0.995 & 0.500 & 0.505 [0.490, 0.524] & \textbf{0.981} \\
SigLIP2-so400m & attn-pool (MAP head) & 0.650 & \textbf{0.937} & $+0.286$ & $+0.020$ & 1.000 & 0.500 & 0.500 [0.485, 0.515] & \textbf{0.990} \\
ConvNeXt-CLIP & global average pool (conv) & 0.937 & \textbf{0.990} & $+0.053$ & $+0.010$ & 1.000 & 0.500 & 0.500 [0.481, 0.519] & \textbf{0.951} \\
RN50-CLIP & attn-pool over conv map & 0.816 & \textbf{0.913} & $+0.097$ & $-0.000$ & 1.000 & 0.500 & 0.471 [0.437, 0.505] & \textbf{0.893} \\
\midrule
\multicolumn{4}{l}{\emph{median}} & $+0.140$ & $-0.002$ & & & & \\
\bottomrule
\end{tabular}
\caption{\textbf{The lateral polarity is in the flip-odd component while the native caption-specific flip effect does not read it.} Left: a probe on $\zm$ beats one on the full pooled vector on every encoder, while the same projection does nothing vertically. Middle: $\zp$ is \emph{exactly} chance on mirror pairs, as Lemma~\ref{lem:odd} requires. Right, both columns under \emph{mirror ranking}, a photograph against its own flip: the caption-specific flip effect is at chance on every encoder, while the dense \moment{} reads the polarity through that same protocol. Cluster bootstraps over object pairs. In every table, \textbf{bold} marks the best value in a group of directly comparable numbers; here, $z$ against $\zm$ and the flip effect against the moment.}\label{tab:main}
\end{table*}

\noindent\textbf{Setup.} What's~Up Controlled~A~\cite{whatsup}: real photographs of object pairs in
\{on, under, left of, right of\}, $n{=}412$, with matched minimal pairs in which only the subject
moves, so size, background and viewpoint are fixed within a pair (Appendix~\ref*{app:size}); and VSR~\cite{vsr}, $1{,}224$
lateral items over $1{,}108$ COCO~\cite{coco} photographs. Ten encoders with a text
tower~\cite{clip,openclip,metaclip,datacomp,dfn,siglip2,convnext} span contrastive and spatially-supervised objectives and CLS, attention-pool/MAP, and global-average-pool heads, two of them convolutional; checkpoints in Tab.~\ref*{tab:encoders}. Every real-image result carries a blind text-only prior and a shuffled-query null, and every lateral number sits beside its vertical counterpart and an object-identity control, per encoder in Appendix~\ref*{app:negative}. Left and
right are the viewer's throughout, in image coordinates. Probes are logistic regressions on
standardised features, with folds grouped by object pair (What's~Up) or image (VSR) and the
regularisation chosen inside the training folds; intervals are cluster bootstraps over the same
groups (Appendices~\ref*{app:protocol} and~\ref*{app:compute}).

\noindent\textbf{Two ranking protocols.} In \emph{photo-pair ranking} the two photographs of one
object pair compete under a fixed caption (Sec.~\ref{sec:bounddef}); in \emph{mirror ranking} a
photograph competes with its own flip, which by Prop.~\ref{prop:tta} is the sign of
$\langle\zm(I), t(c)\rangle$. Neither leaves a free \emph{orientation bit}, a sign convention a scorer could get right without
reading the relation: a per-caption offset cancels
and the caption states which image should win. Two natural alternatives do leave one and manufacture
positive results (Appendix~\ref*{app:calib}).

\noindent\textbf{Probing the flip-odd component.} A probe reads the lateral relation from $\zm$ at
$0.771$--$0.990$ against $0.650$--$0.937$ from the whole pooled vector, so projecting the even
component away \emph{raises} decodability on every encoder (median $\Delta = +0.140$) and changes
nothing vertically (median $-0.002$). Separating an image from its own mirror is much easier: a probe
does it at $\ge 0.990$, while $\zp$ returns exactly $0.500$, as Lemma~\ref{lem:odd} requires.

\noindent\textbf{Native scoring under mirror ranking.} Every mirror test runs with the correct
caption and with the opposite one, and only the difference is a spatial score: on CLIP~ViT-B/16's
vertical axis the raw value is $0.689$ against a wrong-caption $0.660$, almost all of it a preference
for unflipped photographs. The lateral caption-specific effect is $0.471$--$0.510$, every interval
covering $0.5$ (Tab.~\ref{tab:main}); DFN-B, the one encoder clearing a raw one-sided gate at
$0.631$, has a wrong-caption value of $0.621$, and below-chance raw values are the same artifact
reversed (Appendix~\ref*{app:negative}). Photo-pair ranking agrees, a mean $0.502$ over the ten
(Sec.~\ref{sec:bound}). The
protocol is not the limitation: through mirror ranking the \moment{} of Sec.~\ref{sec:method} reads
the polarity at $0.893$--$0.990$.

\noindent\textbf{What's~Up against VSR.} On What's~Up the pooled probe reads
laterality at $0.65$--$0.94$. On VSR it reads $0.475$--$0.527$ against a blind text-only prior of
$0.501$, and not from task difficulty: restricting
VSR to its $56$ frequent category pairs ($n{=}635$) and grouping by pair, the What's~Up protocol
exactly, leaves the pooled probe at $0.481$--$0.542$ while the dense \moment{} reads
$0.723$--$0.754$. Decodability without reachability is established for clean
two-object photographs; on cluttered scenes the pooled vector does not linearly retain the
polarity.

\section{Alignment as the Limiting Factor}
\label{sec:bound}

\begin{table}[t]
\centering\small
\setlength{\tabcolsep}{4pt}
\begin{tabular}{@{}lccc@{}}
\toprule
Encoder & best in $V_{128}$ & real caps. & $\cos(\delta_T, w^\star)$ \\
\midrule
CLIP ViT-B/16 & \textbf{0.690} & 0.524 & $-0.071$ \\
CLIP LAION-2B & \textbf{0.728} & 0.510 & $-0.038$ \\
CLIP DataComp & \textbf{0.835} & 0.490 & $-0.000$ \\
MetaCLIP-B & \textbf{0.709} & 0.558 & $+0.032$ \\
DFN-B & \textbf{0.775} & 0.505 & $+0.042$ \\
SigLIP2-B & \textbf{0.767} & 0.490 & $+0.023$ \\
SigLIP2-L & \textbf{0.718} & 0.510 & $+0.037$ \\
SigLIP2-so400m & \textbf{0.748} & 0.485 & $+0.030$ \\
ConvNeXt-CLIP & \textbf{0.971} & 0.481 & $+0.079$ \\
RN50-CLIP & \textbf{0.932} & 0.510 & $-0.011$ \\
\bottomrule
\end{tabular}
\caption{\textbf{The caption subspace is not the bottleneck; alignment is.} $V_{128}$ is the top-128 principal subspace of $20$K COCO caption embeddings, $\approx\!85\%$ of caption energy. Col.~2: the best bias-free single direction in it, the functional form a caption score has. Col.~3: the best of $22$ relational phrasings, selected on training folds and scored held out. Col.~4: its angle to the image-side lateral direction.}
\label{tab:bound}
\end{table}

Estimating $V$ from $20$K COCO captions (Sec.~\ref{sec:bounddef}), Tab.~\ref{tab:bound} sets the
best bias-free direction inside $V_k$ against the best of $22$ real relational caption pairs,
selected on training folds and scored held out. Inside the subspace holding $85\%$ of caption energy
a caption-shaped direction reaches $0.69$--$0.97$, so the caption \emph{span} is not the constraint;
the phrasings one actually writes reach $0.48$--$0.56$, and their relation direction is
near-orthogonal to the image-side one on every encoder, $|\cos(\delta_T, w^\star)| \le 0.079$. In the
terms of Eq.~\ref{eq:decomp} the encoding is healthy and the alignment fails
(Fig.~\ref*{fig:dissociation}).

\noindent\textbf{A larger search, and where it succeeds.} We search realizable captions directly, $10^6$
on all ten encoders (Appendix~\ref*{app:reach}), selecting each encoder's best pair on a fit half of
the object pairs and scoring the held-out half against a decision level of $0.65$ fixed before the
search. \textbf{Nine remain
near chance} (medians $0.490$--$0.558$; no seed above $0.65$).
\textbf{ConvNeXt-CLIP does not.} There the pool finds a caption pair reading a median $0.678$ held
out over $20$ independent splits (mean $0.664$, minimum $0.567$, above $0.65$ on $14$ of $20$;
subspace bound $0.971$ at $k{=}128$): there a realizable caption reaches the polarity. Its direction is near-orthogonal to $w^\star$ on every split (mean $|\cos| = 0.055$), and a discrete
optimiser validated on a hidden in-family target returns chance on the same split.

That encoder is the positive control for the geometric account. Fed the
direction the search found, Eq.~\ref{eq:predict} gives $0.615$ against $0.515$ for an ordinary
phrasing, a five-fold change in alignment, and ConvNeXt-CLIP's pooled probe reads laterality highest
of the ten ($0.937$ against $0.650$--$0.816$), so it needs the least alignment to clear chance; both
quantities come from the fit half. The subspace bound does not predict reachability across encoders, and
paraphrase and token-optimiser searches on three encoders agree (Appendix~\ref*{app:reach}).

\noindent\textbf{Predicting reachability from geometry.} Under an
equal-covariance model $z \mid y = \pm1 \sim \mathcal{N}(\pm\Delta\mu/2, \Sigma)$, a fixed scoring
direction $\delta$ gives
\begin{equation}
  P(\text{correct}) = \Phi\!\left(\frac{\delta^{\!\top}\Delta\mu}
                                       {2\sqrt{\delta^{\!\top}\Sigma\,\delta}}\right),
  \label{eq:predict}
\end{equation}
so reachability depends on alignment \emph{and} on how much within-class scatter lies along the
scoring direction. Every quantity on the right is measured independently of the accuracy on the
left, which makes it falsifiable: predicted and observed native accuracy differ by a mean absolute
error of $0.016$ laterally and $0.022$ vertically, correlating at $r = 0.950$ on the vertical axis,
the one with real spread across encoders ($0.505$--$0.728$). Evaluated at the Fisher-optimal
direction $\Sigma^{-1}\Delta\mu$ the same expression gives $0.935$ laterally, and the gap to the
observed $0.502$ on identical representations is the reachability deficit, from geometry alone. It
also says what a repair has to change, the alignment term, which neither a larger model nor
patch--text spatial pretraining changes: two TIPS-v2 checkpoints read $0.481$ and $0.476$ laterally
(Appendix~\ref*{app:transport}). Sec.~\ref{sec:supervision} tests relational training.

\section{Pooling Failure on Controlled $C_4$ Orbits}
\label{sec:c4}

\begin{table}[t]
\centering\scriptsize
\setlength{\tabcolsep}{3pt}
\begin{tabular}{@{}lcccccc@{}}
\toprule
& \multicolumn{3}{c}{pooled, binary (chance $0.5$)} & \multicolumn{3}{c}{4-way (chance $0.25$)} \\
\cmidrule(lr){2-4}\cmidrule(lr){5-7}
Encoder & axis & polarity & polarity$^{\dagger}$ & probe & native & moment \\
\midrule
CLIP ViT-B/16 & 1.000 & 0.507 & 0.513 & 0.475 & 0.252 & \textbf{0.819} \\
CLIP LAION-2B & 0.998 & 0.470 & 0.557 & 0.476 & 0.284 & \textbf{0.877} \\
CLIP DataComp & 1.000 & 0.490 & 0.497 & 0.481 & 0.273 & \textbf{0.886} \\
MetaCLIP-B & 1.000 & 0.503 & 0.530 & 0.487 & 0.286 & \textbf{0.823} \\
DFN-B & 1.000 & 0.520 & 0.580 & 0.481 & 0.266 & \textbf{0.858} \\
SigLIP2-B & 1.000 & 0.550 & 0.583 & 0.488 & 0.289 & \textbf{0.967} \\
SigLIP2-L & 1.000 & 0.477 & 0.517 & 0.489 & 0.302 & \textbf{0.962} \\
SigLIP2-so400m & 1.000 & 0.463 & 0.477 & 0.484 & 0.311 & \textbf{0.986} \\
ConvNeXt-CLIP & 1.000 & 0.540 & 0.557 & 0.494 & 0.289 & \textbf{0.951} \\
RN50-CLIP & 1.000 & 0.587 & 0.613 & 0.508 & 0.275 & \textbf{0.811} \\
\midrule
\emph{axis-only ceiling} & \multicolumn{3}{c}{---} & \multicolumn{3}{c}{0.500} \\
\bottomrule
\end{tabular}
\caption{\textbf{Pooling keeps the arrangement axis and loses the polarity along it.} $300$ exact $C_4$ orbits: two segmented objects at a fixed pair centre, displacement rotated $90^\circ$ per step, object pixels upright, so the four members differ only in the relation. \emph{axis} and \emph{polarity} are directions fitted on these orbits and tested on held-out ones. $\dagger$ is the best of an MLP, an RBF-SVM and a cosine kNN on the same rows, orbit groups and folds. On those same rows all three recover an XOR label planted at the polarity channel's own amplitude at $0.978$--$1.000$, which is what makes this column a null rather than an underpowered reader. A reader that identifies the axis and guesses the polarity scores $0.5$ on the 4-way task, which is where the \emph{probe} column sits. The channel is isotropic ($|\cos(u_x,u_y)| \le 0.019$).}\label{tab:c4}
\end{table}

Sections~\ref{sec:measure}--\ref{sec:bound} establish an access failure; this section establishes
a pooling failure, on stimuli where the relation is the only thing that varies.

\noindent\textbf{The construction.} Two segmented COCO objects are composited onto a flat background
at a fixed pair centre, and the displacement between them is rotated by $90^\circ$ per step:
$d_{k+1} = R_{90}\,d_k$, with the object pixels left upright. The four members of an orbit therefore
differ \emph{only} in the relation, which cycles from $A$ right of $B$ to below, then left of, then
above, and $R^4 = \mathrm{id}$ holds by construction. We use above/below, not the on/under of
What's~Up, which carries support and gravity priors and is not the coordinate counterpart of
left/right under this group. $300$ orbits, $1{,}200$ images.

\noindent\textbf{The channels.} The real Fourier decomposition of $C_4$ splits the four embeddings
into an invariant part, a frequency-1 pair
\begin{equation}
  u_x = \tfrac{1}{2}(z_0 - z_2), \quad u_y = \tfrac{1}{2}(z_1 - z_3),
  \label{eq:c4}
\end{equation}
and a frequency-2 part. The flip-odd component of Sec.~\ref{sec:framework} is the $u_x$ half of
this, so $C_4$ contains the $Z_2$ case instead of sitting beside it. The pair is internally
isotropic on every encoder (Tab.~\ref{tab:c4}): one two-dimensional channel, not a horizontal one
with a vertical afterthought.

\noindent\textbf{Axis and polarity, separately.} The pooled vector decodes the 4-way relation at $0.475$--$0.508$ against a chance of $0.25$. Split by what it decides, \emph{which axis} the pair is arranged along is read
at $0.998$--$1.000$,
while the \emph{polarity} along an axis is at chance on nine of the ten: a direction fitted on these
orbits and tested on held-out orbits reads the lateral polarity at $0.463$--$0.587$, and the same
construction through the mirror at $0.480$--$0.562$. We count a direction as reading the polarity
when it clears $0.60$ held out. Of $30$ cells, ten encoders each on the full orbit set, a $207$-orbit
subset, and that subset on a photographic background, one does: RN50-CLIP, the only
attention-pooled convolutional encoder here, reads $0.587$ on the full set and $0.628$
$[0.560, 0.691]$ on the subset (Appendix~\ref*{app:frames}). A line cannot tell a polarity that is gone from one it cannot see, so an
MLP, an RBF-SVM and a cosine kNN repeat the measurement on the same rows, groups and folds. Of $80$
cells, ten encoders by four readers (the MLP at two widths) by two input designs, one clears $0.60$,
RN50-CLIP again ($0.613$ $[0.560, 0.670]$), and the other $79$ span
$0.427$--$0.587$; on those rows all three recover an XOR label planted at the polarity channel's own
amplitude at $0.978$--$1.000$, so this is a statement about the representation and not about the
reader (Appendix~\ref*{app:c4nonlinear}). The $0.475$ is axis identification almost exactly, the
ceiling for a reader that knows the axis and guesses the polarity, and native scoring sits at
$0.252$--$0.311$. The information is not gone from the
image: the zero-parameter \moment{} reads the full 4-way relation from the patch grid at
$0.811$--$0.986$ on the same photographs. \textbf{Pooling, not the encoder, is where the polarity is
lost on these stimuli}: $I_0$ and $I_2$ contain the same two objects at the same two positions and
differ only in which object sits where. On What's~Up the text and image relation planes are
near-orthogonal as whole planes, at every text-tower depth (Appendix~\ref*{app:depth}).

\begin{figure*}[t]
\centering
\includegraphics[width=\textwidth]{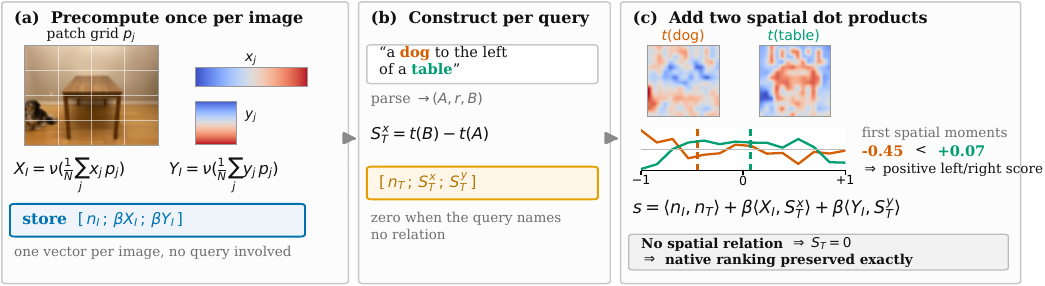}
\caption{\textbf{What is stored, what a query adds, and what their product reads.} In (c), the
fields are centred patch relevance for each object name; dashed lines mark the first spatial moments
used by the score.}
\label{fig:method}
\end{figure*}

\section{A Query-Conditioned Spatial Moment}
\label{sec:method}
\label{sec:bridge}

Both failures sit downstream of the patch grid, which carries the relation in each case
(Tab.~\ref{tab:c4}). Re-reading the pooled vector could at best address the access failure, and
changing the caption side addresses neither (Appendix~\ref*{app:transport}), so the repair recomputes
the geometry from the patch grid, in the form a vector index serves: one vector stored per image, one
computed per query. Its two parts are the moment of Eq.~\ref{eq:moment}, which factorises the readout
into those two vectors, and the pooling-head decomposition of Eq.~\ref{eq:decompose}, which supplies
joint-space patch features for any encoder. The relation word only chooses a sign, so no caption
direction has to align with the image-side one: the repair does not correct the alignment term of
Eq.~\ref{eq:decomp}, it stops depending on it.

\noindent\textbf{The moment.} Patch token $j$ of $N$ sits at grid coordinates
$(x_j, y_j) \in [-1,1]^2$, $x_j$ growing rightward and $y_j$ downward, with feature
$p_j \in \mathbb{R}^D$ in the joint image--text space. Per image we store
\begin{equation}
  X_I = \nu\Big(\tfrac{1}{N}\sum\nolimits_j x_j\, p_j\Big), \qquad
  \langle X_I, t\rangle = c_I \cdot \tfrac{1}{N}\sum\nolimits_j x_j \langle p_j, t\rangle,
  \label{eq:moment}
\end{equation}
and $Y_I$ likewise, with $\nu$ the $L_2$ normalisation and $c_I > 0$ its factor. Up to $c_I$,
$\langle X_I, t\rangle$ is the first \moment{} of the relevance field $\langle p_j, t\rangle$ of any
text embedding $t$: a stored vector times a query vector, the interface retrieval already runs on.

\noindent\textbf{Joint-space patch features.} The $p_j$ must lie in the \emph{joint} image--text
space. Applying the final layer norm and projection to each token does not put them there: the last
attention block mixes tokens, and the resulting per-patch score is anti-correlated with the
relation, in line with the background-over-foreground similarity maps CLIP Surgery~\cite{clipsurgery}
traces to the same block. On CLIP ViT-B/16 a moment built from those vectors reads What's~Up lateral
at $0.131$, where the decomposition below reads $0.772$ on the same images. We decompose the
encoder's own pooling head instead. Every head here computes $z = g(\mu)$, $\mu = \sum_j w_j h_j$,
from pre-pooling features $h_j$ with pooling weights $w_j \ge 0$ summing to one (attention, or $1/N$
for an average) and $g$ the layers after pooling, typically a layer norm and a projection. Set
\begin{equation}
  p_j := g(\mu) + J_g(\mu)\,(h_j - \mu),
  \label{eq:decompose}
\end{equation}
with $J_g$ the Jacobian of $g$. Then $\sum_j w_j p_j = z$ exactly for any $g$, and $p_j = g(h_j)$
exactly whenever $g$ is affine, so this re-reads the encoder's own pooling rather than approximating
it. For a CLS-token ViT it gives MaskCLIP's value-path features~\cite{maskclip}; the affine heads and
the one first-order case, ConvNeXt's average pool, are in Appendix~\ref*{app:protocol}.

\noindent\textbf{The score.} Concatenating,
\begin{equation}
  z_I = [\,n_I \,;\, \beta X_I \,;\, \beta Y_I\,], \quad
  z_T = [\,n_T \,;\, S^x_T \,;\, S^y_T\,],
  \label{eq:score}
\end{equation}
with $n_I, n_T$ the native embeddings. A query parsed as \emph{$A$ left of $B$} sets
$S^x_T = t(B) - t(A)$, $S^y_T = 0$, with $t(\cdot)$ the embedding of ``a photo of a $\cdot$'';
\emph{right of} exchanges $A$ and $B$, and \emph{above}/\emph{below} use $S^y_T$ the same way. The
score is $\langle n_I, n_T\rangle + \beta\langle X_I, S^x_T\rangle + \beta\langle Y_I, S^y_T\rangle$,
with $\beta$ on the image side only so that it \emph{is} the spatial channel's weight. A query naming
no relation has $S_T = 0$ and reproduces the native score exactly for any fixed $\beta$ (to
$3.6 \times 10^{-8}$ in float32, Appendix~\ref*{app:detector}), and since $\|z_I\|^2 = 1 + 2\beta^2$
for every image a cosine index ranks the same way.

\noindent\textbf{$\beta$ without labels.} We set $\beta = \hat\sigma_{\text{nat}} /
\hat\sigma_{\text{sp}}$ from unlabelled score statistics on COCO val2017 captions, which share no
image with the pool where $\beta$ is applied. On What's~Up it costs a median $0.010$ against an
exactly enumerated oracle, at most $0.034$. Across a $16\times$ multiplicative grid the relational
contrast rises monotonically ($0.717$ to $0.910$ mean) while full-pool R@1 on the same queries
peaks at $\hat\beta$ itself, so the unlabelled estimate lands where the two objectives meet, at neither one's optimum. Re-estimating it on CC3M or Open Images moves it by $0.82$--$1.33\times$ and retrieval
accuracy by at most $0.019$, and dropping either object term from $S_T$ costs $0.13$ at every
multiplier (Appendix~\ref*{app:beta}).

\noindent\textbf{Alternative readouts do not explain the gain.} A query-conditioned attention pool
trained on the same $47$K composited items fits that distribution ($0.980$--$1.000$ held out) and
does not transfer, reading $0.447$--$0.592$ on What's~Up in twenty-nine of its thirty trained configurations (ten encoders, $1$, $4$ or $16$ queries); the
zero-parameter rule is above it on all ten encoders, and capacity is not the difference: across
all sixty trained pools (with and without coordinates, $0.8$--$26.3$M parameters), size and out-of-distribution accuracy are unrelated. A relevance-weighted centroid is numerically the \moment{}, while argmax and top-$k$
readers are below it on all four benchmark means. What separates the \moment{} is that it factorises: every
other reader needs the patch grid at query time, $24$--$364\times$ the bytes per image and
$52$--$754\times$ the compute per query (Appendices~\ref*{app:attnpool} and~\ref*{app:densereaders}).

\noindent\textbf{Storage.} A PCA sketch of the spatial channels, fitted on $400$ unlabelled
photographs, costs a median $0.005$ at $k{=}256$ and brings the index to $2\times$ native storage
instead of $3\times$; a zero spatial query stays zero under any linear map, so the sketch keeps the
guarantee (Appendix~\ref*{app:protocol}).

\section{Evaluation at Retrieval Scale}
\label{sec:retrieval}

\begin{table}[t]
\centering\scriptsize
\setlength{\tabcolsep}{2.1pt}
\begin{tabular}{@{}lcccccccc@{}}
\toprule
& \multicolumn{4}{c}{left/right} & \multicolumn{4}{c}{above/below} \\
\cmidrule(lr){2-5}\cmidrule(lr){6-9}
& \multicolumn{2}{c}{pairwise} & \multicolumn{2}{c}{cond.\ R@1} & \multicolumn{2}{c}{pairwise} & \multicolumn{2}{c}{cond.\ R@1} \\
\cmidrule(lr){2-3}\cmidrule(lr){4-5}\cmidrule(lr){6-7}\cmidrule(lr){8-9}
Encoder & nat. & +mom. & nat. & +mom. & nat. & +mom. & nat. & +mom. \\
\midrule
CLIP ViT-B/16 & 0.510 & \textbf{0.839} & 0.048 & \textbf{0.411} & 0.497 & \textbf{0.806} & 0.059 & \textbf{0.312} \\
CLIP LAION-2B & 0.516 & \textbf{0.857} & 0.052 & \textbf{0.451} & 0.527 & \textbf{0.820} & 0.049 & \textbf{0.356} \\
CLIP DataComp & 0.512 & \textbf{0.844} & 0.047 & \textbf{0.421} & 0.523 & \textbf{0.809} & 0.050 & \textbf{0.356} \\
MetaCLIP-B & 0.513 & \textbf{0.808} & 0.046 & \textbf{0.365} & 0.506 & \textbf{0.781} & 0.046 & \textbf{0.289} \\
DFN-B & 0.512 & \textbf{0.837} & 0.045 & \textbf{0.420} & 0.526 & \textbf{0.827} & 0.050 & \textbf{0.343} \\
SigLIP2-B & 0.522 & \textbf{0.938} & 0.051 & \textbf{0.675} & 0.512 & \textbf{0.902} & 0.036 & \textbf{0.520} \\
SigLIP2-L & 0.525 & \textbf{0.952} & 0.059 & \textbf{0.719} & 0.512 & \textbf{0.923} & 0.062 & \textbf{0.570} \\
SigLIP2-so400m & 0.527 & \textbf{0.962} & 0.056 & \textbf{0.753} & 0.524 & \textbf{0.931} & 0.073 & \textbf{0.610} \\
ConvNeXt-CLIP & 0.520 & \textbf{0.862} & 0.051 & \textbf{0.468} & 0.520 & \textbf{0.840} & 0.051 & \textbf{0.440} \\
RN50-CLIP & 0.515 & \textbf{0.838} & 0.049 & \textbf{0.382} & 0.508 & \textbf{0.801} & 0.059 & \textbf{0.314} \\
\midrule
\emph{mean} & 0.517 & \textbf{0.874} & 0.050 & \textbf{0.507} & 0.515 & \textbf{0.844} & 0.053 & \textbf{0.411} \\
\emph{chance} & \multicolumn{2}{c}{0.500} & \multicolumn{2}{c}{0.048} & \multicolumn{2}{c}{0.500} & \multicolumn{2}{c}{0.048} \\
\bottomrule
\end{tabular}
\caption{\textbf{The remedy is not specific to the left/right axis, and it survives as a ranking.} Relational retrieval over 118{,}287 images. Every distractor holds the same two object categories as the gold image with the geometry reversed, so \emph{pairwise} is the fraction of them a gold image outscores and \emph{cond.\ R@1} is how often it is ranked first among all $21$ candidates --- that is, how often the top-ranked image satisfies the queried relation. $n = 6{,}595$ horizontal and $3{,}114$ vertical scored queries; cond.\ R@1 uses the $4{,}812$ and $1{,}110$ with a full pool, so its chance level is a single number. Both scores here are given the parsed $(A, \text{relation}, B)$ triple; free-form queries and the parser are evaluated separately in the supplement. Ambiguous splits and the $\beta$ grid are in the supplement. Cluster bootstraps over query images and over unordered category pairs, whichever is wider, put every pairwise cell inside $\pm0.029$; per-encoder intervals, including the paired difference, are in the supplement.}\label{tab:wp5}
\end{table}

Retrieval is where almost everything in the pool is irrelevant. We build that evaluation from COCO
ground-truth boxes over the $118{,}287$ train2017 images, keeping items where one instance of each of
two categories has unambiguous horizontal separation. Every item carries \textbf{hard negatives holding the same two
categories with the geometry reversed}, which a score that reads objects and ignores arrangement
cannot separate. Labels come from box centres, which a first moment also estimates, so the
box-centre detector below is the matched comparator.

Building it meant looking at the images, not trusting the boxes: a $1\%$ area floor admitted a
$32$-pixel person as ``a person to the left of a bus'', and counting instances above it made ``the
chair'' refer to nothing definite in multi-chair rooms. The clean split shrank from $12{,}248$ to
$6{,}761$ items, of which the $6{,}595$ carrying at least one reversed-geometry distractor are
scored; the $28{,}147$ items failing either test form the separately reported ambiguous split
(Appendix~\ref*{app:ambiguous}).

Tab.~\ref{tab:wp5} is the operational form of the claim, on queries whose
$(A, \text{relation}, B)$ triple is given. \textbf{Native scoring stays close to chance on every
encoder and both axes} ($0.510$--$0.527$ horizontal, $0.497$--$0.527$ vertical), which is what the
construction forces. Queries share images and category pairs, so every cell carries a cluster
bootstrap over both, reported at the wider, with no half-width above $\pm 0.029$. At that resolution
it is not exactly chance: the interval excludes $0.5$ on $12$ of $20$ cells, by $0.010$--$0.027$, the
size of a residual layout prior (Appendix~\ref*{app:wp5ci}). The spatial-moment channel reaches
$0.808$--$0.962$ (mean $0.874$) horizontally and $0.781$--$0.931$ (mean $0.844$) vertically, where
image height partly reflects depth (Appendix~\ref*{app:wp5ci}), at zero learned parameters, a paired difference of $+0.276$ to $+0.435$ whose interval excludes zero
everywhere. The ambiguous splits sit $3$--$5$ points lower (Appendix~\ref*{app:ambiguous}). Query
time over the full pool on CPU is $0.39$--$0.74$\,ms when queries are batched and $3.3$--$8.0$\,ms
for one query issued alone (laptop CPU), after a one-off index build of $21$--$99$ minutes per
encoder (Appendix~\ref*{app:compute}).

\noindent\textbf{Ranking, not only pairing.} Ranking the gold image against all its distractors at
once turns the same pool into a relation-conditioned reranking test: all $21$ candidates already
contain the queried categories, so being ranked first is the event \emph{the top image satisfies the
relation} and nothing else. On the $4{,}812$
horizontal queries with a full pool, conditional R@1 is $0.045$--$0.059$ native against a chance of
$1/21 = 0.048$ and $0.365$--$0.753$ with the moment; vertically $0.289$--$0.610$. Unrestricted
full-pool R@1, where a query naming two categories surfaces them in either arrangement, is a
different and much harder quantity and rises from $0.003$--$0.014$ to $0.013$--$0.037$.

\noindent\textbf{Free-form queries.} Rewriting every query in Qwen2.5-7B-Instruct's own words costs the spatial-moment
score at most $0.003$; recovering the triple from that sentence with a vocabulary-free parser costs the
rest, $0.628$ horizontally against $0.874$ with the triple given and $0.517$ native. The parser
returns a triple for $37\%$ of horizontal queries, read at $0.736$--$0.886$; the rest
fall back to native scoring and account for $87$--$91\%$ of the shortfall. Producing a triple from
ordinary language is a separate problem (Appendix~\ref*{app:freeform}).

\noindent\textbf{Standard caption retrieval.} With the parser in front of every caption of COCO
val2017 ($5{,}000$ images) and Flickr30k ($1{,}000$), it fires on $6.5\%$ and $4.4\%$ of captions,
almost all vertical and some idiomatic (\emph{jumps over}); left/right fires twice per pool. Unparsed
captions score exactly as native. On parsed ones COCO R@1 falls from $0.29$--$0.53$ to
$0.19$--$0.30$: the spatial term has the right sign but, across a whole pool, varies about as much as
the native gap between the gold image and the runner-up. Over all captions R@1 drops by
$0.005$--$0.015$ on every encoder (Appendix~\ref*{app:stdret}). The repair ranks candidates that hold
the named objects; at this $\beta$ it is not a drop-in change to general caption retrieval.

\noindent\textbf{An open-vocabulary detector baseline.} A detector~\cite{owlv2} deciding by box
centres matches the readout in accuracy, both $0.990$ on What's~Up and $0.775$ against $0.763$ on
VSR at comparable coverage. It abstains on items it cannot ground, and must re-run per (image,
query) pair or store boxes and labels at $460$\,MB, neither of which a vector index serves
(Appendix~\ref*{app:detector}).

\section{Controlled Supervision Experiments}
\label{sec:supervision}

Frozen checkpoints invite the objection that relational training would solve the problem by itself.
Fine-tuning CLIP ViT-B/32 on synthetic composites batched with their left/right swaps lifts
VG-Relations left/right from $0.49$ to $0.56$--$0.57$ under image-to-caption choice~\cite{whatsleft},
and small encoders trained from scratch on a one-dimensional synthetic testbed learn the
relation~\cite{lrsymmetry}. We test ViT-B/32 CLIP models on natural images, scored on What's~Up under
the paired protocol of Sec.~\ref{sec:measure}. Twenty-four of them, trained from scratch on $373{,}528$
annotated images ($29.06$M samples each; eight caption conditions, three seeds), with the fraction
$p$ of captions stating a true relation running to $1.0$, read native lateral accuracy
$0.5004 \pm 0.0176$ with no trend in $p$ ($r = -0.23$). Supervision moves the wrong term of Eq.~\ref{eq:decomp}: the text contrast does not
develop, $\cos(t(\text{left}), t(\text{right}))$ staying at $0.987$ at $p = 1$, inside the public
encoders' $0.972$--$0.996$, since a contrastive loss never has to separate two words used in
near-identical contexts. Varying only phrasing over $2{,}100$ constructions
moves reachability by at most $+0.008$, and making the captions false half the time is
indistinguishable. Each anchor's exact left/right counterfactual in its batch, the ingredient
of~\cite{whatsleft}, reads $0.471$--$0.529$ over three seeds against $0.398$--$0.485$ without it,
inside the null region ($\le 0.55$) of a rule fixed before training. Three fine-tunes of a
pretrained ViT-B/32 (LoRA, NegCLIP-style negatives, in-batch swaps on our own renders) stay inside
$0.485$--$0.515$ at $256$k pairs while the \moment{} reads $0.70$--$0.76$ before and after.
\textbf{The supervision strategies and scales tested here do not restore native reachability}, which
does not contradict~\cite{whatsleft,lrsymmetry}: models, data and protocols differ (Appendices~\ref*{app:finetune}, \ref*{app:rho},
\ref*{app:supmech}, \ref*{app:axis2}; Tab.~\ref*{tab:wp6}).

\section{Limitations}
\label{sec:limits}

The diagnosis exploits the horizontal-flip and $C_4$ symmetries, so it is established for left/right
and above/below; relations without that structure, front/behind among them, are outside what these
measurements can decide. Caption reachability is estimated from a finite set of realizable
captions, and on one encoder of the ten a $10^6$-caption search does reach the polarity, the
positive control of Sec.~\ref{sec:bound}. The \moment{} assumes a parsed $(A, \text{relation}, B)$ query and
unambiguous object references, and on standard caption benchmarks it lowers R@1 on the captions
the parser fires on (Sec.~\ref{sec:retrieval}). As a first moment it gives each named object one
location and cannot resolve several instances of a category, which is where the ambiguous split
loses $0.03$--$0.04$ (Appendix~\ref*{app:ambiguous}). It also needs the pre-pooling features and the
encoder's pooling head, so unlike the flip test-time augmentation diagnostic it cannot be applied to
an API returning only a final pooled vector; its two spatial channels take the index to $3\times$
native storage, or about $2\times$ with the sketch of Sec.~\ref{sec:method}.

\section{Conclusion}

A near-chance spatial similarity score does not identify one failure. On clean two-object
photographs the lateral polarity survives pooling, decodable from the flip-odd part at
$0.771$--$0.990$, but realizable captions point near-orthogonally to it, so native scoring cannot
read it. On controlled $C_4$ orbits pooling keeps the arrangement axis at $0.998$--$1.000$ and, on
nine of ten encoders, drops the polarity along it, which the patch grid still carries. One
zero-parameter readout of that grid covers both: over $118{,}287$ images it takes left/right
pairwise accuracy from $0.51$--$0.53$ to $0.81$--$0.96$ and conditional R@1 among $21$
category-matched candidates from $0.05$ to $0.37$--$0.75$, at under a millisecond per batched query,
leaving spatially empty rankings unchanged. Relational supervision, in the forms and at the scales we
tested, does not restore it. Open are what sets the alignment, why pooling discards the polarity,
whether front/behind splits the same way, object-centred frames (the chair's own left), and orbits
that rotate the objects or move them in depth.

\noindent\textbf{Societal impact.} The study collects no new human data and uses public benchmarks
and public checkpoints, whose vocabulary and scenes the readout inherits, meaning COCO categories,
English relation words and two unambiguous objects.

{\small
\bibliographystyle{ieeenat_fullname}
\bibliography{bib/main}
}

\maketitlesupplementary

\noindent Tables, figures and equations numbered with an \textbf{S} prefix are in this
document; every other number refers to the main paper.

\appendix
\setcounter{table}{0}
\setcounter{figure}{0}
\setcounter{equation}{0}
\renewcommand{\thetable}{S\arabic{table}}
\renewcommand{\thefigure}{S\arabic{figure}}
\renewcommand{\theequation}{S\arabic{equation}}
\renewcommand{\theHtable}{S\arabic{table}}
\renewcommand{\theHfigure}{S\arabic{figure}}
\renewcommand{\theHequation}{S\arabic{equation}}
\section{Protocol}
\label{app:protocol}

\begin{table}[t]\centering\scriptsize\setlength{\tabcolsep}{4pt}
\begin{tabular}{@{}ll@{}}
\toprule
Encoder & checkpoint \\
\midrule
CLIP ViT-B/16 & \texttt{ViT-B-16 / openai} \\
CLIP LAION-2B & \texttt{ViT-B-16 / laion2b\_s34b\_b88k} \\
CLIP DataComp & \texttt{ViT-B-16 / datacomp\_xl\_s13b\_b90k} \\
MetaCLIP-B & \texttt{ViT-B-16-quickgelu / metaclip\_fullcc} \\
DFN-B & \texttt{ViT-B-16 / dfn2b} \\
SigLIP2-B & \texttt{google/siglip2-base-patch16-224} \\
SigLIP2-L & \texttt{google/siglip2-large-patch16-256} \\
SigLIP2-so400m & \texttt{google/siglip2-so400m-patch14-384} \\
ConvNeXt-CLIP & \texttt{convnext\_base\_w / laion2b\_s13b\_b82k} \\
RN50-CLIP & \texttt{RN50 / openai} \\
\midrule
\multicolumn{2}{@{}l}{\emph{transport only} (Appendix~\ref{app:transport})} \\
CLIP ViT-B/32 LAION-2B & \texttt{ViT-B-32 / laion2b\_s34b\_b79k} \\
TIPS-v2-B & \texttt{google/tipsv2-b14} \\
TIPS-v2-so400m & \texttt{google/tipsv2-so400m14} \\
\bottomrule
\end{tabular}
\caption{\textbf{The evaluated checkpoints.} The ten encoders with a text tower used throughout, plus the three appearing only in the symmetry-transport analysis; pooling heads are in Tab.~\ref*{tab:main}. A \texttt{model / tag} pair is an OpenCLIP architecture and pretrained tag, a single string a Hugging Face identifier.}\label{tab:encoders}
\end{table}

\noindent\textbf{Encoders.} Tab.~\ref{tab:encoders} lists every checkpoint. All are used frozen, at their own native
resolution and preprocessing.

\noindent\textbf{Benchmarks.} What's~Up Controlled~A~\cite{whatsup}: $412$ photographs, $206$ of them
lateral, over $103$ object pairs with matched minimal pairs. VSR~\cite{vsr} random split, restricted
to lateral relations: $1{,}224$ items over $1{,}108$ COCO~\cite{coco} images. Captions for VSR are
templated as ``The $X$ is \{left,right\} of the $Y$'' from the released subject/object parse.

\noindent\textbf{Probes.} Logistic regression on standardised features, group $k$-fold with groups
formed by object pair (What's~Up) or image (VSR), and the regularisation constant chosen by an inner
group $k$-fold \emph{on the training folds only}. Reported accuracy therefore never sees the value
that selected its hyperparameter. Intervals are cluster bootstraps resampling groups, not items,
because several items share an image.

\noindent\textbf{Paired ranking.} For an object pair with both photographs present, and for each of
the two captions, the true image is ranked against the false one. A per-caption offset cancels
identically, and the caption states which image should win, so no orientation bit exists.

\noindent\textbf{Densification.} Eq.~\ref*{eq:decompose} re-reads the encoder's own pooling head.
Two further properties: the mean-centred query moment depends on $t$ only through
$J_g(\mu)^{\!\top} t$, so the constant $g(\mu)$ drops out; and MaskCLIP's value-projection
bypass~\cite{maskclip}, the MAP head applied per token, and RN50's attention pool are all instances
with affine $g$, while ConvNeXt's global average pool is the one first-order case.

\begin{table}[t]
\centering\small
\setlength{\tabcolsep}{4pt}
\begin{tabular}{@{}lccccc@{}}
\toprule
& & \multicolumn{2}{c}{PCA sketch} & \multicolumn{2}{c}{random projection} \\
\cmidrule(lr){3-4}\cmidrule(lr){5-6}
$k$ & storage & acc. & cost & acc. & cost \\
\midrule
16 & 1.03--1.06$\times$ & 0.748 & $+0.090$ & 0.598 & $+0.212$ \\
32 & 1.06--1.12$\times$ & 0.774 & $+0.056$ & 0.683 & $+0.137$ \\
64 & 1.11--1.25$\times$ & 0.784 & $+0.036$ & 0.696 & $+0.113$ \\
128 & 1.22--1.50$\times$ & 0.808 & $+0.019$ & 0.718 & $+0.078$ \\
256 & 1.44--2.00$\times$ & 0.820 & $+0.005$ & 0.751 & $+0.043$ \\
\midrule
full & $3\times$ & 0.833 & --- & --- & --- \\
native & $1\times$ & 0.502 & --- & --- & --- \\
\bottomrule
\end{tabular}
\caption{\textbf{What the compressed index costs, over the whole rank range.} Median across 10 encoders on What's~Up lateral; storage is stored floats per image as a multiple of the native vector. \emph{cost} is the drop from the uncompressed $3D$ representation, so a negative entry beats it. PCA is fitted on $400$ unlabelled photographs available at index-build time; the random projection is the assumption-free construction, reported because it is the one that fails.}\label{tab:sketch}
\end{table}

\noindent\textbf{Storage, across the rank range.} Tab.~\ref{tab:sketch} is the whole curve rather
than the single operating point quoted in Sec.~\ref*{sec:retrieval}. A random projection of the
spatial channels is not good enough at any rank --- a median $0.751$ even at $k{=}256$, where it
still distorts the score by a median $51\%$ against the PCA sketch's $27\%$ --- and it is reported
because it is the construction
that carries the distortion bound. The PCA sketch of Sec.~\ref*{sec:method} is what the paper uses;
its cost falls monotonically in $k$, from $0.090$ at $k{=}16$ to $0.005$ at $k{=}256$. The basis is fitted on $400$ unlabelled photographs disjoint from everything scored, so
the sketch stays label-free. One caveat travels with it: the spatial channel needs $72$ dimensions
for $95\%$ of its energy and $136$ for $99\%$, estimated from $206$ items, so that rank is bounded
by $n$ and is not a population estimate.

\section{Calibration and Its Failure Modes}
\label{app:calib}

One scoring variant scores $\langle \zm, t(X) - t(Y)\rangle$ using \emph{object-name} directions. That
contrast is oriented by object identity, so a genuine coordinate convention remains and one bit must
be fixed. We fix it on $16$ synthetic renders whose layout we control, never on a benchmark, and
report agreement against a \emph{second, independent} render set built from different shapes,
colours, backgrounds, scales and separations.

That bit is unstable on $3$ of $10$ encoders (agreement as low as $0.56$ on $16$ items), which is why
this variant is not the primary score anywhere, and why the mirror-ranking protocol, which has
no convention bit at all, carries the paper.

No such bit exists for caption-oriented contrasts, and applying one would be selecting orientation on
the evaluation set. Below-chance values are reported as below chance.

\noindent\textbf{Two protocols that manufacture positive results.} The paired direction used in
Sec.~\ref*{sec:measure} was adopted after both of these were tried. First, caption-selection
accuracy, which scores an image against its two candidate captions, is dominated by a class-independent
offset: per-class splits such as $0.99/0.07$ occur, so the mean estimates a near-constant predictor
rather than a discrimination, and only the split reveals it. Second, applying a global sign to a
\emph{label-oriented} score $t_{\text{correct}} - t_{\text{opposite}}$ turns any below-chance value
into an above-chance one; because the label has already oriented the contrast, no convention bit
remains to calibrate, and doing so selects orientation on the evaluation set. The paired protocol
has neither problem.

\section{Controls and Negative Results}
\label{app:negative}

\begin{table}[t]
\centering\scriptsize
\setlength{\tabcolsep}{3.5pt}
\begin{tabular}{@{}lcccc@{}}
\toprule
& obj.-identity & wrong-cap. & VSR blind & \multicolumn{1}{c}{\moment{}: real / shuf.} \\
Encoder & variant & null & prior & (What's~Up lat.) \\
\midrule
CLIP ViT-B/16 & 0.563 & 0.510 & 0.501 & 0.772 / 0.588 \\
CLIP LAION-2B & 0.597 & 0.451 & 0.501 & 0.806 / 0.635 \\
CLIP DataComp & 0.563 & 0.398 & 0.501 & 0.694 / 0.523 \\
MetaCLIP-B & 0.549 & 0.393 & 0.501 & 0.772 / 0.626 \\
DFN-B & 0.437 & 0.621 & 0.501 & 0.782 / 0.630 \\
SigLIP2-B & 0.500 & 0.374 & 0.501 & 0.859 / 0.751 \\
SigLIP2-L & 0.612 & 0.500 & 0.501 & 0.966 / 0.741 \\
SigLIP2-so400m & 0.534 & 0.350 & 0.501 & 0.990 / 0.875 \\
ConvNeXt-CLIP & 0.583 & 0.456 & 0.501 & 0.893 / 0.749 \\
RN50-CLIP & 0.553 & 0.481 & 0.501 & 0.743 / 0.528 \\
\bottomrule
\end{tabular}
\caption{\textbf{The priors and nulls behind every real-image number.} Under photo-pair and mirror ranking both competing photographs carry the \emph{same} caption, so a text-only reader is exactly $0.500$ by construction rather than by measurement; VSR is unpaired, so its blind prior is measured (col.~4). Col.~2 scores the lateral relation from object-name directions instead of the relation, the object-identity control. Col.~3 is the shuffled-query null for native scoring: the same mirror test run with the opposite caption, whose difference from the raw value is the flip effect of Tab.~\ref*{tab:main}. Col.~5 is the shuffled-query null for the \moment{}, its score when given \emph{another item's} object names; the real score is above it on every encoder, by $0.108$--$0.225$.}
\label{tab:controls}
\end{table}

\noindent\textbf{Which null belongs to which measurement.} Sec.~\ref*{sec:measure} and
Sec.~\ref*{sec:related} state that every real-image number carries a blind text-only prior and a
shuffled-query null, and that every lateral number sits beside a vertical counterpart and an
object-identity control. Tab.~\ref{tab:controls} prints them per encoder. The blind prior is not one
quantity: under photo-pair and mirror ranking the two competing photographs carry the same caption,
so a text-only reader scores exactly $0.500$ by construction, and the same holds at retrieval scale,
where every distractor holds the queried categories; only VSR, which has no reversed-geometry
partner, needs a measured one. The shuffled-query null is likewise the strongest available form of
itself in each case: for native scoring it is the same mirror test under the \emph{opposite}
caption, a targeted swap rather than a random one, and Tab.~\ref*{tab:main}'s flip effect is already
the difference between the two; for the \moment{} it is another item's object names. The vertical
counterpart is $\Delta_{\text{vert}}$ in Tab.~\ref*{tab:main} and the $y$ axis throughout.

Each item below changed a number we would otherwise have reported.

\noindent\textbf{Naturalness inflates every raw flip test.} An inverted photograph is bizarre in a
way a mirrored one is not, so ``upright beats flipped'' can hold for any caption. On CLIP
ViT-B/16's vertical axis, $0.660$ of an apparent $0.689$ is present with the \emph{opposite}
caption. Every mirror test is
reported as raw / wrong-caption / difference.

\noindent\textbf{Below-chance raw values.} Several encoders sit far below chance on the raw number (SigLIP2-so400m lateral is the lowest at $0.350$), with intervals excluding $0.5$. A one-sided gate would file those as null when a consistent inverted effect would be a reachable signal of reversed polarity. The wrong-caption control resolves it: that encoder scores $0.350$ with the correct caption \emph{and} $0.350$ with the opposite one, so the value is a caption-independent preference for the mirrored image and carries no relational information.

\noindent\textbf{Per-class splits before means.} Caption-selection accuracies here include splits of
$0.99/0.07$ by class: near-constant predictors whose means estimate nothing. We check the split
before interpreting any mean.

\noindent\textbf{Two exactly-$0.500$ results that were bugs.} The bias-free caption bound first
returned precisely $0.500$ at every $k$, including full rank where nothing is restricted, which reads
as a sweeping impossibility theorem. First cause: both orderings of each pair had been enumerated and
labelled positive, making the problem unlearnable. Second, found only because one encoder still did
it: these towers are near mirror-invariant, so $\|\zm\| \approx 0.05$ and at a fixed regularisation
strength the penalty drives the solution to zero. The bound then becomes regularisation-limited
instead of information-limited, by a per-encoder amount. Normalising each difference to unit norm
(the decision is invariant to positive rescaling) and selecting the constant on inner folds fixes it. Every
encoder's value moved.

\noindent\textbf{CCA reads shared image structure.} Canonical correlation between $\zm$ and the
lateral \moment{} is $0.95$--$0.98$ at $p = 0.000$; the \emph{vertical} \moment{} scores as high or
higher. 

\noindent\textbf{The oracle must be computed exactly.} Accuracy is a step function of $\beta$ whose
breakpoints are $-s_{\text{nat},i}/s_{\text{sp},i}$, so a grid oracle can only land on its own
points while a label-free estimate is continuous, which flatters the label-free arm. We enumerate
breakpoints instead.

\noindent\textbf{A prediction that did not come true.} We predicted that global average pooling,
whose pooling weights are position-independent, would fail the flip score where position-dependent
heads might pass. It does fail, and so do the position-dependent heads. The predicted dissociation
is absent and the lateral failure is uniform across pooling mechanisms.

\section{Compute and Reporting}
\label{app:compute}

\noindent\textbf{Reporting.} Training runs report three seeds per cell and are quoted per seed, never as mean $\pm$ sd, wherever
a cell sits near a threshold.

Each from-scratch run sees
$29{,}059{,}540$ samples, ten epochs of $2{,}905{,}954$ (the size of CC3M) with a checkpoint per
epoch, at batch size $1{,}024$. That budget is the same for every from-scratch run, including those
trained on the $373{,}528$-image annotated pool of Tab.~\ref{tab:wp6}, which is therefore cycled about
$7.8$ times per epoch. All cells within an axis use the same image corpus, optimisation budget and
number of samples seen.

\noindent\textbf{Optimisation, in full.} Every training run in the paper, as it was run. Values not
listed are the \texttt{open\_clip} defaults for the architecture, and nothing was tuned per cell:
within an axis the arms differ in the corpus or the batch composition and in nothing else.

\begin{table}[!ht]
\centering\scriptsize
\setlength{\tabcolsep}{3pt}
\resizebox{\columnwidth}{!}{%
\begin{tabular}{@{}llll@{}}
\toprule
 & from scratch ($24$ $+$ $6$ runs) & fine-tune ($3$) & learned pool ($60$) \\
\midrule
init & ViT-B/32, random & ViT-B/32, \texttt{laion2b\_s34b\_b79k} & frozen enc.\ $+$ head \\
optimiser & AdamW $(0.9, 0.98)$, $\epsilon{=}10^{-6}$ & AdamW & AdamW \\
lr & $10^{-3}$, cosine & $10^{-4}$ LoRA, $10^{-6}$ full & $3{\times}10^{-4}$, constant \\
warmup & $2{,}000$ steps & none & none \\
weight decay & $0.2$ & $0.1$ & $10^{-2}$ \\
batch & $1{,}024$ & $128$ & $32$ \\
precision & \texttt{amp\_bf16} & \texttt{bf16} autocast & fp32 \\
budget & $10$ ep.\ $\times$ $2{,}905{,}954$ & $2{,}000$ steps ($256$K pairs) & $20$ ep.\ $\times$ $47$K \\
grad.\ clip & none & none & none \\
seeds & $3$ per cell & $1$ & $1$ (fixed) \\
\bottomrule
\end{tabular}}
\caption{Optimisation settings for every training run in the paper, as run. Values not listed are the \texttt{open\_clip} defaults; nothing was tuned per cell.}
\label{tab:optim}
\end{table}

\noindent\textbf{Where the latency and storage numbers were measured.} The query figure is the
scoring step and nothing else: the queries against the $118{,}287$ cached image blocks, as a float32
matrix product in NumPy on a compute node's CPUs (six cores), with no GPU involved. Encoding the
query text is excluded from it, as it is common to the native baseline and to any retrieval system.
That is $0.39$--$0.74$\,ms per query by encoder on the clean split, about a minute for the whole of
it. The figure is wall clock on a shared node and moves with the node: the same code measured
$0.26$--$0.53$\,ms on an earlier run and $0.471$--$0.735$\,ms in the controlled comparison below, and
one contended job reached $4.7$\,ms. Sub-millisecond is the claim; a specific decimal is not.
\textbf{It is a batched figure}: one matrix product covers the whole query set and the time is
divided by the number of queries, which is how a retrieval evaluation runs. A single query issued on
its own is a matrix--vector product instead, which BLAS executes far less efficiently. Measured
both ways on one machine, a laptop CPU, the scoring step costs
$0.21$--$0.39$\,ms per query batched and $3.3$--$8.0$\,ms per query issued alone, across
$D \in \{512, 768, 1152\}$ at the full $118{,}287$-image pool. Both are the same computation per
query; only the batching differs.

Building the index is the expensive half and happens once: $21$--$99$ minutes per encoder on one
GPU, $20.5$--$103.5$ images/s over the same images. On disk the blocks are
float16 --- $115.5$\,MB at $D{=}512$ and $231.0$\,MB at $D{=}1024$ for the native block, three times
that for the uncompressed augmented representation and twice with the $k{=}256$ sketch.

\noindent\textbf{The detector, on the same machine.} The comparison in Appendix~\ref{app:detector}
first put a laptop measurement beside a cluster one, so we re-ran the detector on the node that
produced the query figure. On that node's GPU it costs $0.272$\,s/image on
What's~Up and $0.164$\,s/image on VSR, with its accuracy reproducing ($0.990$ exactly, and $0.776$
against $0.775$ on VSR), while the readout's scoring step on the same node's CPUs is
$0.471$--$0.735$\,ms/query across $D \in \{512, 768, 1152\}$. The gap is what the interface implies:
the detector re-runs a network per image, and the readout multiplies vectors that were stored once.

\noindent Weight decay follows \texttt{open\_clip}'s split: gains, biases and the logit scale are
excluded. The counterfactual arms use the same settings as the from-scratch ones, on a corpus of
$65{,}536$ pairs cycled to the same samples seen, with the sample-shuffle buffer disabled so the
baked batches survive the loader (Appendix~\ref{app:rho}).

\section{The larger caption search}
\label{app:reach}

\begin{figure*}[t]
\centering
\includegraphics[width=\textwidth]{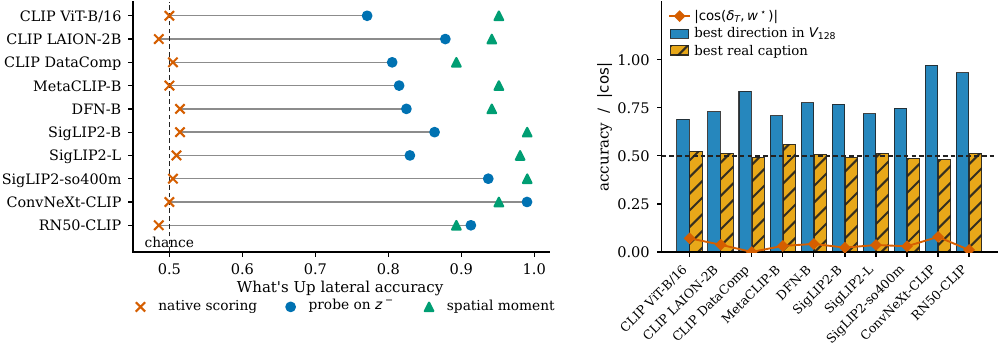}
\caption{\textbf{The decodability--reachability gap under the standard caption set, and the geometry
behind it.} Left: native scoring is at chance on all ten while a probe on the flip-odd component
reads the lateral polarity; the connector is the gap between what the representation holds and what
the metric uses. Right: the best bias-free direction in the caption \emph{span} $V_{128}$ is far
above chance while the best of the $22$ relational caption pairs remains near chance, and the two
relation directions are near-orthogonal.}
\label{fig:dissociation}
\end{figure*}

\noindent\textbf{Span versus realizable set.} A subspace containing a usable direction does not imply
a \emph{caption} producing one: the realizable set $\mathcal{M}_T = \{t(c)\}$ is a subset of that span
and not a subspace of it, so we search it. Crossing $20$ relational frames with orderings,
determiners, prefixes and anchors on real object pairs gives $50{,}400$ captions and $25{,}200$
realizable decision directions per encoder. Across all ten the best reaches
$|\cos(\delta, w^\star)| = 0.042$--$0.201$ while $w^\star$ lies entirely within their span
($\|P_V w^\star\| = 1.000$): inside the span of language, off the surface it occupies.

\noindent\textbf{Larger and smarter searches.} Would a larger or smarter search find what that
$50{,}400$-caption pool does not? We ran three arms on
CLIP ViT-B/16, SigLIP2-so400m and ConvNeXt-CLIP: a $10^6$-caption template pool, $\approx 50$K pairs
from an instruction-tuned paraphrase model (Qwen2.5-7B-Instruct~\cite{qwen25}), and a greedy discrete optimiser over token slots. The
$103$ What's~Up lateral object pairs are split in half by pair; $w^\star$ is fitted on the fit half
only, each arm selects its single best candidate by fit accuracy alone, and the number reported is
that one direction's accuracy on the held half, which never enters selection.

\noindent\textbf{The optimiser arm, and the gate it needed.} Its first capability gate --- recover a
hidden but reachable target direction to $|\cos| \ge 0.90$ --- failed on all three encoders
($0.29$--$0.69$), for a structural reason: the hidden target was a comparative construction
(``further to the left than'') and the optimiser's slot family emits only prepositional ones, so the
search space could not represent it. A second gate, its target drawn inside the slot family and fixed
by a disjoint seed before the run, is recovered at $|\cos| = 1.000$ on all three. The procedure works;
the first gate measured a mismatch between a target's grammar and the family being searched. With the
gate passing, the optimiser reads chance everywhere: $0.500$--$0.519$ held out under a caption
skeleton, and $0.500$ with the skeleton dropped, on every encoder tried.

\noindent\textbf{Ten encoders, one of them reachable.} Nine read chance held out --- medians
$0.490$--$0.558$ over five splits each, no seed anywhere above $0.65$, and $|\cos(\delta, w^\star)|$
between $0.011$ and $0.074$ --- against subspace bounds of $0.737$--$0.941$, each at its own best
$k$ in the sweep. ConvNeXt-CLIP reads
$0.683$ from the $10^6$-caption pool and $0.596$ from the paraphrase model, against a subspace bound
of $0.990$, which is $0.971$ at the $k{=}128$ Tab.~\ref*{tab:bound} reports. The winning pair is
\emph{``somewhat a picture of the the blanket lies on the left-hand side of the the chair''}
against its right-hand counterpart --- a string that survives selection, not one anybody would type.

\noindent\textbf{It replicates.} Repeating the whole $10^6$-caption search under $20$ independent
fit/held splits gives a median held-out $0.678$, mean $0.664$, standard deviation $0.043$, minimum
$0.567$, and $14$ of $20$ splits above $0.65$. The first split was not a lucky draw, and no split
returns chance. Two facts hold across all twenty. The direction is near-orthogonal to $w^\star$ on
every one of them (mean $|\cos| = 0.055$, range $0.017$--$0.096$), so what the caption reads is not
the lateral channel the rest of the paper measures. And restricting the candidate pool to object
names drawn from the fit half alone returns the identical winning caption and the identical
accuracy, so nothing here comes from naming a held-out pair.

\noindent\textbf{A validated optimiser cannot find it either.} On ConvNeXt-CLIP, on the same
fit/held split where enumeration over $10^6$ candidates returns $0.678$, greedy coordinate search
from a random start returns exactly $0.500$ --- with its gate passing on that encoder. Two readings
fit: the successful region is narrow enough that local search misses it, or the two searches reach
different parts of the realizable caption set. We do not adjudicate between them; either way the pool's
result is a point rather than a neighbourhood.

\noindent\textbf{The paper's own model predicts it.} Eq.~\ref*{eq:predict} turns geometry into an
accuracy, and ConvNeXt-CLIP is the encoder with the most encoding to work with: a probe on its
pooled vector reads laterality at $0.937$ against $0.650$--$0.816$ for the other nine, so the
alignment a caption needs to clear chance is smallest there. Fitting $\Delta\mu$ and $\Sigma$ on the
fit half alone and feeding in the direction the search actually found, the predictor gives $0.615$
on the held half, against $0.635$ observed for that fixed direction and the $0.683$ the search
reports for its sign rule. The same model, same encoder, same geometry, fed an ordinary relational
phrasing instead, gives $0.515$. What separates them is alignment: $\cos(\delta, \Delta\mu)$ is
$0.128$ for the found caption and $0.023$ for the ordinary one. So this result sits inside the
account of Sec.~\ref*{sec:framework} rather than beside it --- on the encoder where the encoding term
is largest, a five-fold improvement in alignment is enough, and a caption search over $10^6$
candidates finds one.

\noindent\textbf{Does reachability track the bound?} Across all ten encoders held-out accuracy and
the subspace bound correlate at $r = 0.675$, which is one high-leverage point: dropping
ConvNeXt-CLIP leaves $r = 0.190$ (Spearman $\rho = 0.412$, $p = 0.27$), and RN50-CLIP has the
second-highest bound and reads exactly chance. The bound does not predict reachability on this
evidence.

\noindent\textbf{What this changes.} On ConvNeXt-CLIP the lateral polarity is reachable by a
realizable caption, and the paper says so. It is one encoder of ten searched at this scale; the
other nine stay at chance. The claim is about the encoders it was measured on rather than about
language, which is the form a finite search can support.

\section{Simpler readers off the same patch grid}
\label{app:densereaders}

\begin{table}[t]
\centering\small
\setlength{\tabcolsep}{4pt}
\begin{tabular}{@{}lcccc@{}}
\toprule
Reader & WU l/r & WU on/under & VSR & $C_4$ 4-way \\
& \emph{(0.5)} & \emph{(0.5)} & \emph{(0.5)} & \emph{(0.25)} \\
\midrule
\moment{} (ours) & 0.826 & 0.662 & \textbf{0.738} & 0.894 \\
centroid, $T{=}1$ & 0.827 & \textbf{0.664} & \textbf{0.738} & \textbf{0.895} \\
centroid, $T{=}4$ & \textbf{0.828} & 0.663 & \textbf{0.738} & \textbf{0.895} \\
centroid, $T{=}16$ & 0.827 & 0.662 & \textbf{0.738} & \textbf{0.894} \\
argmax patch ($k{=}1$) & 0.713 & 0.602 & 0.666 & 0.490 \\
top-$k$ mean, $k{=}4$ & 0.753 & 0.622 & 0.713 & 0.657 \\
top-$k$ mean, $k{=}16$ & 0.760 & 0.614 & 0.733 & 0.767 \\
top-$k$ mean, $k{=}$ all & 0.500 & 0.500 & 0.501 & 0.250 \\
\bottomrule
\end{tabular}
\caption{\textbf{Simpler readers off the same patch grid, and what they cost.} Mean over 10 encoders; chance in italics. Every row reads the same dense features and differs only in how it turns a relevance field into a position. The centroid family is the \moment{} numerically, and $k{=}$ all is chance by construction. What separates the \moment{} is not accuracy but that it factorises: every other row needs the patch grid at query time, $24$--$364\times$ the bytes and $52$--$754\times$ the compute.}\label{tab:densereaders}
\end{table}

Eq.~\ref*{eq:moment} turns a relevance field into a position by weighting every patch. Simpler rules
do the same job, and if one of them matched at the same cost the construction would not be earning
its place. We compare the family that reads the identical dense features and differs only in that
step: the relevance-weighted centroid through a softmax at $T \in \{1, 4, 16\}$, the argmax patch,
and the mean coordinate of the $k$ most relevant patches for $k \in \{1, 4, 16, \text{all}\}$.

\noindent\textbf{Three benchmarks of four separate them.} The \moment{} ties the centroid family,
which is numerically the same rule, and is above every other reader on all four benchmark means. Per encoder the lead is uneven: it holds on all ten for the
$C_4$ orbits, on eight for What's~Up lateral and on five for VSR, and where it loses it loses by at
most $0.011$ on VSR and $0.058$ on What's~Up lateral. The gap is widest on $C_4$
($0.894$ against $0.767$ for the best alternative), where a single most-relevant patch is a poor
estimate of an orientation that the whole field carries.

\noindent\textbf{The fourth splits by encoder, and informatively.} On What's~Up above/below, the six
encoders where the \moment{} sits below $0.60$ are exactly the six where an alternative beats it, by
$0.01$--$0.14$; the four where it is above $0.75$ keep the lead by $0.03$--$0.18$. Simpler readers do
not have a general edge --- they recover part of what the \moment{} misses on its weakest axis for
the encoders where it is weakest there.

\noindent\textbf{Two boundary cases that look like results.} The centroid at $T = 1$, $4$ and $16$ is
numerically the \moment{} (differences $\le 0.003$ everywhere): relevance here is a cosine similarity
with a spread of roughly $0.2$--$0.4$, so even $T{=}1$ already sits in the near-uniform regime whose
limit the \moment{} is. And $k = \text{all}$ is exactly chance on every benchmark and every encoder,
because an unweighted mean over the whole grid returns the same coordinate for both named objects.
Neither is evidence about the family; both are what its own algebra requires.

\noindent\textbf{What actually separates them.} Not accuracy but form. Every other reader here needs
the patch grid present at query time --- $24$--$364\times$ the bytes stored per image and
$52$--$754\times$ the compute per query, measured on one machine --- while
$\langle X_I, t\rangle$ is a dot product between something stored once and something computed per
query. A reader that matches the \moment{} and does not factorise cannot be served from an index,
which is the interface the retrieval claim is about.

\section{Uncertainty for the retrieval tables}
\label{app:wp5ci}

\noindent\textbf{The vertical axis, and a disjoint pool.} Pooling drops the polarity on either axis
(Sec.~\ref*{sec:c4}), and the readout recovers both at scale. On natural photographs that axis partly
conflates height with depth, which is why the $C_4$ orbits are its controlled counterpart. A
replication on a disjoint pool (COCO val2017, $1{,}199$ images) agrees to within $0.06$ for every encoder, axis and score.

The retrieval queries are not independent draws, so an interval built as though they were would be
too narrow. The clean horizontal split holds $6{,}761$ items, of which the $6{,}595$ with at least
one reversed-geometry distractor are scored. Those sit on $4{,}553$ images --- up
to ten queries share one photograph --- and use $831$ unordered category pairs, one of which carries
$120$ queries. Queries that share a photograph, or share a pair of categories, succeed and fail
together.

Image clusters and category-pair clusters are crossed rather than nested: a photograph carries
several pairs and a pair spans many photographs, so neither one-way resampling contains the other.
We therefore bootstrap both, $2{,}000$ resamples each, and report the wider, which needs no
assumption about how the two dependencies interact. Because both arms are scored on the same
queries, the paired difference is also resampled, and that is the interval the comparison rests on.
The same resampling is applied to the $\beta$ grid and to the one-term ablations.

\begin{table*}[t]
\centering\scriptsize
\setlength{\tabcolsep}{4pt}
\begin{tabular}{@{}lcccccc@{}}
\toprule
& \multicolumn{3}{c}{left/right} & \multicolumn{3}{c}{above/below} \\
\cmidrule(lr){2-4}\cmidrule(lr){5-7}
Encoder & native & +moment & $\Delta$ & native & +moment & $\Delta$ \\
\midrule
CLIP ViT-B/16 & 0.510 [0.501, 0.519] & 0.839 [0.826, 0.851] & 0.329 [0.314, 0.344] & 0.497 [0.475, 0.520] & 0.806 [0.784, 0.826] & 0.309 [0.286, 0.329] \\
CLIP LAION-2B & 0.516 [0.507, 0.525] & 0.857 [0.845, 0.868] & 0.341 [0.327, 0.354] & 0.527 [0.502, 0.552] & 0.820 [0.797, 0.841] & 0.293 [0.267, 0.318] \\
CLIP DataComp & 0.512 [0.503, 0.522] & 0.844 [0.830, 0.856] & 0.332 [0.317, 0.346] & 0.523 [0.498, 0.547] & 0.809 [0.788, 0.829] & 0.286 [0.263, 0.311] \\
MetaCLIP-B & 0.513 [0.504, 0.523] & 0.808 [0.793, 0.822] & 0.294 [0.279, 0.310] & 0.506 [0.481, 0.531] & 0.781 [0.758, 0.803] & 0.276 [0.253, 0.299] \\
DFN-B & 0.512 [0.503, 0.521] & 0.837 [0.823, 0.851] & 0.325 [0.310, 0.340] & 0.526 [0.502, 0.550] & 0.827 [0.810, 0.844] & 0.302 [0.278, 0.325] \\
SigLIP2-B & 0.522 [0.513, 0.532] & 0.938 [0.930, 0.945] & 0.416 [0.405, 0.426] & 0.512 [0.489, 0.537] & 0.902 [0.886, 0.917] & 0.390 [0.366, 0.414] \\
SigLIP2-L & 0.525 [0.515, 0.534] & 0.952 [0.946, 0.958] & 0.427 [0.417, 0.437] & 0.512 [0.483, 0.541] & 0.923 [0.910, 0.934] & 0.411 [0.385, 0.437] \\
SigLIP2-so400m & 0.527 [0.518, 0.536] & 0.962 [0.957, 0.967] & 0.435 [0.425, 0.445] & 0.524 [0.495, 0.552] & 0.931 [0.920, 0.941] & 0.407 [0.382, 0.433] \\
ConvNeXt-CLIP & 0.520 [0.510, 0.530] & 0.862 [0.849, 0.873] & 0.342 [0.328, 0.355] & 0.520 [0.496, 0.544] & 0.840 [0.820, 0.860] & 0.320 [0.294, 0.345] \\
RN50-CLIP & 0.515 [0.507, 0.525] & 0.838 [0.825, 0.851] & 0.323 [0.310, 0.336] & 0.508 [0.486, 0.531] & 0.801 [0.781, 0.822] & 0.293 [0.272, 0.314] \\
\bottomrule
\end{tabular}
\caption{\textbf{Cluster-robust intervals for every cell of the clean retrieval split.} $95\%$ bootstrap, $2{,}000$ resamples. The queries are not independent draws --- the 6{,}595 scored horizontal ones sit on 4{,}553 images and use 831 unordered category pairs --- so we resample clusters, by image and by category pair, and report the wider. Both arms score the same queries, so $\Delta$ is a paired difference. Widest half-width here: $\pm0.029$.}\label{tab:wp5ci}
\end{table*}

\begin{table*}[t]
\centering\scriptsize
\setlength{\tabcolsep}{4pt}
\begin{tabular}{@{}lcccccc@{}}
\toprule
& \multicolumn{3}{c}{left/right} & \multicolumn{3}{c}{above/below} \\
\cmidrule(lr){2-4}\cmidrule(lr){5-7}
Encoder & native & +moment & $\Delta$ & native & +moment & $\Delta$ \\
\midrule
CLIP ViT-B/16 & 0.498 [0.494, 0.503] & 0.807 [0.797, 0.817] & 0.308 [0.299, 0.318] & 0.504 [0.495, 0.513] & 0.783 [0.770, 0.795] & 0.279 [0.267, 0.291] \\
CLIP LAION-2B & 0.497 [0.492, 0.501] & 0.810 [0.801, 0.820] & 0.314 [0.304, 0.323] & 0.508 [0.498, 0.518] & 0.792 [0.780, 0.804] & 0.284 [0.271, 0.297] \\
CLIP DataComp & 0.498 [0.493, 0.502] & 0.803 [0.793, 0.813] & 0.306 [0.296, 0.315] & 0.510 [0.501, 0.520] & 0.789 [0.776, 0.801] & 0.278 [0.265, 0.291] \\
MetaCLIP-B & 0.497 [0.492, 0.501] & 0.762 [0.752, 0.772] & 0.265 [0.256, 0.275] & 0.508 [0.498, 0.518] & 0.750 [0.737, 0.762] & 0.242 [0.230, 0.254] \\
DFN-B & 0.498 [0.493, 0.502] & 0.799 [0.789, 0.810] & 0.302 [0.291, 0.311] & 0.513 [0.504, 0.522] & 0.788 [0.776, 0.800] & 0.276 [0.264, 0.288] \\
SigLIP2-B & 0.495 [0.490, 0.499] & 0.893 [0.886, 0.901] & 0.399 [0.391, 0.406] & 0.504 [0.494, 0.515] & 0.866 [0.855, 0.877] & 0.362 [0.349, 0.374] \\
SigLIP2-L & 0.493 [0.488, 0.498] & 0.912 [0.905, 0.918] & 0.418 [0.412, 0.425] & 0.509 [0.498, 0.519] & 0.886 [0.875, 0.897] & 0.377 [0.366, 0.389] \\
SigLIP2-so400m & 0.493 [0.488, 0.497] & 0.924 [0.918, 0.929] & 0.431 [0.424, 0.437] & 0.512 [0.501, 0.523] & 0.898 [0.888, 0.906] & 0.386 [0.374, 0.397] \\
ConvNeXt-CLIP & 0.496 [0.491, 0.500] & 0.816 [0.805, 0.827] & 0.320 [0.310, 0.331] & 0.508 [0.498, 0.518] & 0.812 [0.798, 0.826] & 0.304 [0.290, 0.319] \\
RN50-CLIP & 0.498 [0.493, 0.503] & 0.791 [0.781, 0.802] & 0.293 [0.283, 0.304] & 0.505 [0.495, 0.514] & 0.767 [0.755, 0.779] & 0.263 [0.250, 0.275] \\
\bottomrule
\end{tabular}
\caption{\textbf{Cluster-robust intervals for every cell of the ambiguous retrieval split.} $95\%$ bootstrap, $2{,}000$ resamples. The queries are not independent draws --- the 27{,}551 scored horizontal ones sit on 18{,}210 images and use 1{,}073 unordered category pairs --- so we resample clusters, by image and by category pair, and report the wider. Both arms score the same queries, so $\Delta$ is a paired difference. Widest half-width here: $\pm0.014$.}\label{tab:wp5ciamb}
\end{table*}

\noindent\textbf{Relation-conditioned R@1.} The pairwise cell averages over distractors. Ranking the
gold image against all of them at once turns the same pool into a retrieval result, and because every
candidate in it already holds both queried categories, being ranked first is the event that the
top-ranked image satisfies the relation. Two details decide whether that number is trustworthy. A tie
does not count as a win: the gold image must score strictly above all $20$ distractors, which is the
comparison the evaluator makes. And pool size varies --- $73\%$ of horizontal queries and $36\%$ of
vertical ones carry the full $20$ --- which would otherwise mix chance levels within one column, so
the table reports the full-pool queries, where chance is exactly $1/21$. Over every query regardless
of pool size the chance level is the mean of $1/(n_i+1)$, namely $0.079$; native scoring reads
$0.080$--$0.095$ there and the moment $0.420$--$0.781$, so neither restriction changes the reading.

\begin{table*}[t]
\centering\scriptsize
\setlength{\tabcolsep}{4pt}
\begin{tabular}{@{}lcccccc@{}}
\toprule
& \multicolumn{3}{c}{left/right} & \multicolumn{3}{c}{above/below} \\
\cmidrule(lr){2-4}\cmidrule(lr){5-7}
Encoder & native & +moment & $\Delta$ & native & +moment & $\Delta$ \\
\midrule
CLIP ViT-B/16 & 0.048 [0.042, 0.055] & 0.411 [0.378, 0.443] & 0.362 [0.330, 0.395] & 0.059 [0.043, 0.076] & 0.312 [0.256, 0.367] & 0.253 [0.202, 0.309] \\
CLIP LAION-2B & 0.052 [0.045, 0.059] & 0.451 [0.414, 0.485] & 0.399 [0.363, 0.435] & 0.049 [0.033, 0.067] & 0.356 [0.299, 0.413] & 0.307 [0.253, 0.362] \\
CLIP DataComp & 0.047 [0.041, 0.053] & 0.421 [0.383, 0.458] & 0.374 [0.336, 0.412] & 0.050 [0.030, 0.072] & 0.356 [0.299, 0.414] & 0.306 [0.253, 0.360] \\
MetaCLIP-B & 0.046 [0.040, 0.053] & 0.365 [0.331, 0.399] & 0.319 [0.284, 0.353] & 0.046 [0.029, 0.065] & 0.289 [0.240, 0.340] & 0.243 [0.200, 0.288] \\
DFN-B & 0.045 [0.038, 0.051] & 0.420 [0.385, 0.453] & 0.376 [0.342, 0.408] & 0.050 [0.035, 0.066] & 0.343 [0.286, 0.400] & 0.294 [0.236, 0.352] \\
SigLIP2-B & 0.051 [0.045, 0.058] & 0.675 [0.639, 0.708] & 0.624 [0.587, 0.657] & 0.036 [0.024, 0.051] & 0.520 [0.462, 0.580] & 0.484 [0.425, 0.542] \\
SigLIP2-L & 0.059 [0.052, 0.066] & 0.719 [0.688, 0.748] & 0.661 [0.628, 0.691] & 0.062 [0.045, 0.084] & 0.570 [0.511, 0.629] & 0.508 [0.449, 0.567] \\
SigLIP2-so400m & 0.056 [0.049, 0.063] & 0.753 [0.722, 0.781] & 0.697 [0.667, 0.725] & 0.073 [0.050, 0.100] & 0.610 [0.554, 0.666] & 0.537 [0.479, 0.593] \\
ConvNeXt-CLIP & 0.051 [0.045, 0.058] & 0.468 [0.432, 0.505] & 0.417 [0.380, 0.454] & 0.051 [0.036, 0.070] & 0.440 [0.377, 0.500] & 0.388 [0.324, 0.452] \\
RN50-CLIP & 0.049 [0.042, 0.056] & 0.382 [0.349, 0.416] & 0.334 [0.299, 0.368] & 0.059 [0.037, 0.084] & 0.314 [0.258, 0.367] & 0.254 [0.201, 0.305] \\
\bottomrule
\end{tabular}
\caption{\textbf{Relation-conditioned R@1, with the same cluster-robust intervals as the pairwise cells.} How often the gold image is ranked first among the $21$ candidates that all hold the two queried categories. Chance is exactly $1/21 = 0.048$, and native scoring sits on it. Restricted to the 4{,}812 horizontal and 1{,}110 vertical queries whose pool is full, so one chance level covers every row; the mixed-pool variant is in the text. A tie does not count as a win: the readout must score the gold image strictly above all $20$.}\label{tab:wp5condr1}
\end{table*}

\section{Sensitivity to $\beta$ and to the object terms}
\label{app:beta}

$\beta$ is one scalar, set from unlabelled score statistics rather than fitted, so the question is
how much it can be wrong. The grid is multiplicative around the label-free $\hat\beta$ and spans a
factor of $16$. The same table drops one term of the query-side contrast $S_T = t(B) - t(A)$: keeping
only $t(Y)$, or only $-t(X)$, at the same $\hat\beta$ and, in the result files, over the same
multiplier grid.

\begin{table}[t]
\centering\scriptsize
\setlength{\tabcolsep}{3pt}
\begin{tabular}{@{}lcccccccc@{}}
\toprule
& \multicolumn{5}{c}{$\beta$ multiplier} & \multicolumn{2}{c}{one-term $S_T$} & \\
\cmidrule(lr){2-6}\cmidrule(lr){7-8}
Encoder & $\times0.25$ & $\times0.5$ & $\times1$ & $\times2$ & $\times4$ & $B$-only & $A$-only & native \\
\midrule
CLIP ViT-B/16 & 0.677 & 0.768 & 0.839 & 0.872 & 0.883 & 0.695 & 0.697 & 0.510 \\
CLIP LAION-2B & 0.704 & 0.793 & 0.857 & 0.885 & 0.892 & 0.723 & 0.733 & 0.516 \\
CLIP DataComp & 0.677 & 0.771 & 0.844 & 0.879 & 0.890 & 0.708 & 0.721 & 0.512 \\
MetaCLIP-B & 0.648 & 0.731 & 0.808 & 0.853 & 0.868 & 0.650 & 0.655 & 0.513 \\
DFN-B & 0.678 & 0.769 & 0.837 & 0.870 & 0.880 & 0.713 & 0.722 & 0.512 \\
SigLIP2-B & 0.776 & 0.879 & 0.938 & 0.958 & 0.963 & 0.828 & 0.821 & 0.522 \\
SigLIP2-L & 0.796 & 0.897 & 0.952 & 0.969 & 0.973 & 0.846 & 0.839 & 0.525 \\
SigLIP2-so400m & 0.818 & 0.916 & 0.962 & 0.975 & 0.979 & 0.856 & 0.849 & 0.527 \\
ConvNeXt-CLIP & 0.709 & 0.799 & 0.862 & 0.889 & 0.897 & 0.745 & 0.741 & 0.520 \\
RN50-CLIP & 0.690 & 0.776 & 0.838 & 0.864 & 0.872 & 0.695 & 0.702 & 0.515 \\
\midrule
\emph{mean} & 0.717 & 0.810 & 0.874 & 0.901 & 0.910 & 0.746 & 0.748 & 0.517 \\
\emph{mean R@1} & 0.019 & 0.022 & 0.024 & 0.023 & 0.021 & --- & --- & 0.008 \\
\bottomrule
\end{tabular}
\caption{\textbf{The repair does not need $\beta$ to be right, and it does need both object terms.} Pairwise accuracy, clean horizontal split. Multipliers apply to the label-free $\hat\beta$, so $\times 1$ is the printed setting and the columns span a factor of $16$. \emph{$B$-only} and \emph{$A$-only} keep one term of $S_T = t(B) - t(A)$; no multiplier rescues either. The last row is why $\hat\beta$ is not simply too small: R@1 peaks there while the relational contrast keeps rising.}\label{tab:wp5beta}
\end{table}

\noindent\textbf{Calibrating somewhere else.} $\hat\beta$ is estimated on COCO val2017, which is the
same corpus family as the index, so Tab.~\ref{tab:betaood} re-estimates it by the identical
unlabelled rule on CC3M and on Open Images photographs. Every value lands within
$0.82$--$1.33\times$ the COCO one, and the spread between corpora is no larger than the spread
between $400$-image subsamples of a single corpus. What settles it is applying each estimate
unchanged to the retrieval evaluation: over $40$ comparisons --- ten encoders, two corpora, both
axes --- the repaired arm's accuracy moves by at most $0.019$, with a mean change of $+0.0009$. The calibration corpus is not
something the method is tuned on.

\begin{table}[t]
\centering\small
\setlength{\tabcolsep}{5pt}
\begin{tabular}{@{}lccccc@{}}
\toprule
& COCO & \multicolumn{2}{c}{CC3M} & \multicolumn{2}{c}{Open Images} \\
\cmidrule(lr){3-4}\cmidrule(lr){5-6}
Encoder & $\hat\beta$ & $\hat\beta$ & ratio & $\hat\beta$ & ratio \\
\midrule
CLIP ViT-B/16 & 0.79 & 0.75 & $0.95\times$ & 0.68 & $0.86\times$ \\
CLIP LAION-2B & 1.30 & 1.18 & $0.91\times$ & 1.07 & $0.82\times$ \\
CLIP DataComp & 1.01 & 1.05 & $1.04\times$ & 0.99 & $0.98\times$ \\
MetaCLIP-B & 0.79 & 0.83 & $1.05\times$ & 0.70 & $0.88\times$ \\
DFN-B & 1.03 & 1.19 & $1.16\times$ & 1.01 & $0.99\times$ \\
SigLIP2-B & 0.74 & 0.95 & $1.28\times$ & 0.82 & $1.11\times$ \\
SigLIP2-L & 0.79 & 1.01 & $1.27\times$ & 0.87 & $1.10\times$ \\
SigLIP2-so400m & 0.84 & 1.11 & $1.33\times$ & 0.87 & $1.04\times$ \\
ConvNeXt-CLIP & 0.96 & 0.98 & $1.02\times$ & 0.95 & $0.99\times$ \\
RN50-CLIP & 0.80 & 0.89 & $1.12\times$ & 0.82 & $1.03\times$ \\
\bottomrule
\end{tabular}
\caption{\textbf{The calibration corpus does not matter.} $\hat\beta$ re-estimated by the same unlabelled rule on two corpora that are not COCO. Every value is within $0.82$--$1.33\times$ the COCO one, no further than five disjoint $400$-image subsamples of a single corpus move it ($1.06$--$1.28\times$). Open Images ships no captions, so its caption side is CC3M alt-text.}\label{tab:betaood}
\end{table}

\section{Free-form queries, end to end}
\label{app:freeform}

\begin{table}[t]
\centering\small
\setlength{\tabcolsep}{3.5pt}
\begin{tabular}{@{}lcccccc@{}}
\toprule
& \multicolumn{3}{c}{left/right} & \multicolumn{3}{c}{above/below} \\
\cmidrule(lr){2-4}\cmidrule(lr){5-7}
Encoder & templ. & free & parsed & templ. & free & parsed \\
\midrule
CLIP ViT-B/16 & 0.839 & 0.839 & 0.610 & 0.806 & 0.806 & 0.663 \\
CLIP LAION-2B & 0.857 & 0.857 & 0.616 & 0.820 & 0.823 & 0.675 \\
CLIP DataComp & 0.844 & 0.846 & 0.611 & 0.809 & 0.807 & 0.665 \\
MetaCLIP-B & 0.808 & 0.811 & 0.602 & 0.781 & 0.784 & 0.647 \\
DFN-B & 0.837 & 0.840 & 0.616 & 0.827 & 0.828 & 0.666 \\
SigLIP2-B & 0.938 & 0.939 & 0.657 & 0.902 & 0.903 & 0.717 \\
SigLIP2-L & 0.952 & 0.952 & 0.660 & 0.923 & 0.923 & 0.725 \\
SigLIP2-so400m & 0.962 & 0.962 & 0.667 & 0.931 & 0.931 & 0.741 \\
ConvNeXt-CLIP & 0.862 & 0.863 & 0.625 & 0.840 & 0.842 & 0.694 \\
RN50-CLIP & 0.838 & 0.839 & 0.614 & 0.801 & 0.804 & 0.669 \\
\midrule
\emph{mean} & 0.874 & 0.875 & 0.628 & 0.844 & 0.845 & 0.686 \\
\emph{native} & \multicolumn{3}{c}{0.517} & \multicolumn{3}{c}{0.515} \\
\bottomrule
\end{tabular}
\caption{\textbf{A free-form query costs nothing; recovering the triple from it costs everything.} Three arms on the clean split. \emph{templ.} is the printed protocol. \emph{free} replaces the caption with one Qwen2.5-7B-Instruct wrote in its own words, objects still given, and differs by $-0.0014$ to $+0.0034$ across all twenty cells. \emph{parsed} takes the triple from a vocabulary-free parser run on that sentence, with unparsed queries scored natively rather than dropped.}\label{tab:freeform}
\end{table}

The retrieval protocol hands the readout a triple. A deployed search box does not, so we put the
whole chain in front of it and measured the two costs separately. An instruction-tuned model, Qwen2.5-7B-Instruct~\cite{qwen25} sampled at temperature $0.8$ with at most $40$ new tokens, rewrote
each templated query in its own words, with an automatic check that both objects and a relation of
the correct polarity survive; $6{,}252$ of $6{,}761$ horizontal and $3{,}215$ of $3{,}658$ vertical
captions passed it. A vocabulary-free parser then recovered $(A, \text{relation}, B)$ from that
sentence, and queries it could not parse were scored natively rather than dropped.

\noindent\textbf{The caption channel is free.} Swapping a template for a sentence a person would type
moves the repaired arm by $-0.0014$ to $+0.0034$ across all twenty cells. Whatever the score reads,
it does not need the phrasing to be canonical.

\noindent\textbf{Coverage is the larger term.} End to end the mean falls from $0.874$ to $0.628$
horizontally and $0.844$ to $0.686$ vertically, still far above the native $0.517$ and $0.515$. The
parser returns a triple for $37.1\%$ of horizontal queries and $80.7\%$ of vertical ones, and a query
it cannot parse falls back to native scoring. Splitting the gap between the free-form arm and the
end-to-end one across the three groups a query can fall into, the unparsed group accounts for
$87$--$91\%$ of it horizontally and $39$--$44\%$ vertically.

\noindent\textbf{A name that differs from the ground-truth string is not a failed parse.} Requiring
the parsed name to match the COCO category exactly makes ``a dining table'' $\to$ ``table'' an error,
and on that measure the parser names both objects correctly for $9.1\%$ of horizontal and $2.1\%$ of
vertical queries. What the readout consumes is a text embedding, not a string. Grouped by the string
test, queries whose parsed names differ from the ground truth are read at $0.706$--$0.858$
horizontally against $0.828$--$0.969$ for an exact match, both far above the native $0.517$; the
vertical axis gives $0.682$--$0.796$ against $0.720$--$0.895$. The string test is not what governs
the chain.

\noindent What this locates is where the interface work is. The readout consumes a triple; producing
one from ordinary language is a parsing problem, and the parser here is deliberately simple, with no
object vocabulary. Raising its coverage is what the third column would pay for.

\section{Ambiguous Retrieval Splits}
\label{app:ambiguous}

Tab.~\ref{tab:wp5-ambiguous} reports the pairwise contrast on the pre-declared ambiguous splits, kept
separate from the clean results throughout.

\begin{table}[h]
\centering\footnotesize
\begin{tabular}{@{}lcccc@{}}
\toprule
& \multicolumn{2}{c}{left/right} & \multicolumn{2}{c}{above/below} \\
\cmidrule(lr){2-3}\cmidrule(lr){4-5}
Encoder & native & +moment & native & +moment \\
\midrule
CLIP ViT-B/16 & 0.498 & \textbf{0.807} & 0.504 & \textbf{0.783} \\
CLIP LAION-2B & 0.497 & \textbf{0.810} & 0.508 & \textbf{0.792} \\
CLIP DataComp & 0.498 & \textbf{0.803} & 0.510 & \textbf{0.789} \\
MetaCLIP-B & 0.497 & \textbf{0.762} & 0.508 & \textbf{0.750} \\
DFN-B & 0.498 & \textbf{0.799} & 0.513 & \textbf{0.788} \\
SigLIP2-B & 0.495 & \textbf{0.893} & 0.504 & \textbf{0.866} \\
SigLIP2-L & 0.493 & \textbf{0.912} & 0.509 & \textbf{0.886} \\
SigLIP2-so400m & 0.493 & \textbf{0.924} & 0.512 & \textbf{0.898} \\
ConvNeXt-CLIP & 0.496 & \textbf{0.816} & 0.508 & \textbf{0.812} \\
RN50-CLIP & 0.498 & \textbf{0.791} & 0.505 & \textbf{0.767} \\
\midrule
\emph{mean} & 0.496 & 0.832 & 0.508 & 0.813 \\
\bottomrule
\end{tabular}
\caption{Pairwise retrieval on the pre-declared ambiguous splits, kept separate from the headline clean results. Horizontal: 28{,}147 queries; vertical: 22{,}782.}
\label{tab:wp5-ambiguous}
\end{table}

\section{Relational fine-tuning}
\label{app:finetune}

\begin{table}[t]
\centering\small
\setlength{\tabcolsep}{4pt}
\begin{tabular}{@{}lcccc@{}}
\toprule
& trainable & WU zero-shot & VSR & ViewSpatial \\
& params & lateral & moment & moment \\
\midrule
base, none & --- & 0.505 & 0.712 & \textbf{0.755} \\
LoRA & 0.86\,M & 0.515 & 0.699 & 0.722 \\
NegCLIP & 151\,M & 0.495 & 0.712 & 0.712 \\
synthetic & 151\,M & 0.485 & \textbf{0.717} & 0.736 \\
\bottomrule
\end{tabular}
\caption{\textbf{Relational fine-tuning does not move the native score.} All three start from the same ViT-B/32 checkpoint and are compared against \emph{that} model; 256k supervised pairs, two of them full fine-tunes. Zero-shot lateral stays inside 0.485--0.515 while the zero-parameter \moment{} reads 0.70--0.76 before and after. At 2000 steps the claim is about this scale, not about supervision in general.}\label{tab:wp4}
\end{table}

Three fine-tunes of one ViT-B/32 checkpoint (\texttt{laion2b\_s34b\_b79k}), $2000$ steps, $256$k
supervised pairs, batch $128$: a LoRA adapter at $0.86$\,M trainable parameters, a NegCLIP-style run
with compositional hard negatives, and a synthetic-render arm in the style of~\cite{whatsleft}.
Two of the three are full fine-tunes at $151$\,M parameters. The third trains on our own MuJoCo
renders with that paper's objective, not on its data or its exact recipe, neither of which is
publicly obtainable; it is evidence about training on synthetic scenes at this budget and is not a
reproduction of their reported number.

\noindent\textbf{The control matters more than the arms.} All three start from the same checkpoint,
and their numbers are uninterpretable without \emph{that} model's own: ``LoRA reaches $0.515$
lateral'' says nothing until the base is known to score $0.505$. The base is \emph{not} the
\texttt{clip-laion2b} row of Tab.~\ref*{tab:main}, which is a different architecture
(ViT-B/16) and a different checkpoint. We verified that it loads $151{,}277{,}313$
parameters, matching the total recorded in all three fine-tune runs.

\noindent\textbf{What moves.} Nothing operational. What's~Up native lateral scoring stays inside $0.485$--$0.515$ against chance $0.500$; the VSR native probe is $0.458$--$0.459$ against a base of
$0.458$, unchanged to three decimals and below the blind prior throughout. The \moment{} readout
reads $0.70$--$0.76$ before and after, and is if anything marginally \emph{lower} after fine-tuning.

\noindent\textbf{The synthetic arm is the sharpest case.} Its training loss fell to $0.013$, so it
fitted the renders almost perfectly, and transferred none of it to any real benchmark. Training
produced a decodable-but-unreachable representation on demand rather than repairing reachability.

\noindent\textbf{Not a supervised ceiling.} $2000$ steps and $256$k pairs is a modest budget. The
defensible claim is that relational fine-tuning \emph{at this scale} does not move the native score, not that no amount of supervision could.

\section{The detector baseline in full}
\label{app:detector}

\begin{table}[t]
\centering\small
\setlength{\tabcolsep}{3.5pt}
\begin{tabular}{@{}llcc@{}}
\toprule
Bench. & Method & Cov. & Accuracy \\
\midrule
WU & det.\ $\tau{=}0.02$ & 1.000 & \textbf{0.990} [0.971, 1.000] \\
 & det.\ $\tau{=}0.05$ & 0.990 & \textbf{0.990} [0.971, 1.000] \\
 & det.\ $\tau{=}0.1$ & 0.961 & \textbf{0.990} [0.970, 1.000] \\
 & det.\ $\tau{=}0.2$ & 0.883 & 0.989 [0.967, 1.000] \\
 & det.\ $\tau{=}0.3$ & 0.786 & 0.988 [0.963, 1.000] \\
 & moment & 1.000 & \textbf{0.990} [0.971, 1.000] \\
\midrule
VSR & det.\ $\tau{=}0.02$ & 0.992 & 0.775 [0.752, 0.799] \\
 & det.\ $\tau{=}0.05$ & 0.975 & 0.774 [0.750, 0.799] \\
 & det.\ $\tau{=}0.1$ & 0.935 & 0.776 [0.752, 0.801] \\
 & det.\ $\tau{=}0.2$ & 0.731 & 0.791 [0.762, 0.817] \\
 & det.\ $\tau{=}0.3$ & 0.467 & \textbf{0.797} [0.759, 0.830] \\
 & moment & 1.000 & 0.763 [0.738, 0.786] \\
\bottomrule
\end{tabular}
\caption{\textbf{An open-vocabulary detector matches the readout in accuracy.} It abstains when it cannot ground a named object, so it is reported as a coverage/accuracy curve over its threshold $\tau$, from one detection pass; the readout always answers. On VSR its higher numbers are bought with coverage. What separates them is that it cannot be precomputed into an index.}
\label{tab:detector}
\end{table}

The detector is OWLv2~\cite{owlv2} prompted with the two object names, deciding by the $x$-centres
of the highest-scoring box per name. Its score threshold controls coverage and accuracy together, so
a single hand-picked value is not a fair presentation of an abstaining method against one that
always answers: detection runs once at a low threshold and every reported threshold is derived from
the same boxes.

\noindent\textbf{What's~Up is flat.} Accuracy is $0.988$--$0.990$ across thresholds $0.02$--$0.30$
while coverage falls from $1.000$ to $0.786$, so the comparison is not an artefact of the operating
point, and at its best point the detector is $0.990$ at \emph{full} coverage against the
\moment{}'s $0.990$.

\noindent\textbf{VSR is where abstention shows.} The detector reaches $0.797$ only at $0.467$
coverage. At $0.992$ coverage it is $0.775$ against the readout's $0.763$ at $1.000$, a gap of
$0.012$ with overlapping intervals.

\noindent\textbf{Unchanged-ranking measurement.} In float32 the norm spread across images is
$\sim\!10^{-7}$ and empty-query scores agree with native to $3.6 \times 10^{-8}$. A bare equality
test on the induced ordering nevertheless fails on one encoder, because two images there have
exactly equal native scores and the tie is broken arbitrarily; every discordant position is such a
tie, so we report the score agreement, which is unambiguous, next to a tie-aware ordering check. The
retrieval index is float16 for storage, where exact ties are common ($0.20$--$0.31$ per query
against $\sim\!0$ in float32); that index carries R@K and the pairwise contrast, and the
unchanged-ranking evidence is quoted from float32 throughout.

\noindent\textbf{Scoring on the detector's answered and refused subsets.} On siglip2-l: on the
$1{,}194$ items the detector answered, the \moment{} reads $0.764$ against the detector's $0.774$;
on the $30$ it refused, $0.733$ against no answer at all.
The readout performs near its overall rate on the items the detector cannot handle, so those items
are not intrinsically hard. The $n{=}30$ interval is wide ($[0.552, 0.879]$) and the reading is
``consistent with its overall rate, clearly not chance'', not a point estimate.

\noindent\textbf{Cost.} $460$\,MB of additional model, $0.27$--$0.85$\,s per image, and a
query-conditioned interface: a detector must either re-run per (image, query) pair or store boxes
and labels, neither of which a vector index serves. The spatial-moment channel is a dot product against a
stored vector, from the encoder pass retrieval already performs.

\section{Counterfactual support}
\label{app:rho}

Contrastive training could fail to represent a relation for two different reasons: the captions
never mention it, or they mention it while the objective never needs it. The specificity axis rules
out the first. This experiment tests the second.

\noindent\textbf{Definition.} For an anchor, let $\rho$ be the probability that its batch contains
an example whose non-relational content is identical and whose relation differs. The specificity
parameter $p$ controls $P(\text{caption states a relation})$; $\rho$ controls whether the loss can
tell two examples apart \emph{without} using the relation. Note $\rho$ here is anchor support: with
one counterpart per sample and batch size $1024$, full support means every anchor has its
counterpart present, i.e.\ one exact counterfactual among $1023$ negatives.

\noindent\textbf{Construction.} Counterfactuals are exact: the same image horizontally mirrored,
with the relation word swapped and every other word fixed. Batch assignment is baked offline into
shard order and the sample-shuffle buffer is disabled, so the loader delivers exactly the intended
batches; a counterfactual \emph{displaces} an ordinary negative rather than enlarging the batch, so
batch size is constant. At $\rho{=}0$ the two members of a pair are forced into different batches,
making accidental support impossible rather than merely unlikely. Both arms train on the same
$65{,}536$ samples, verified by a hash of the sample keys, so the arms differ in batch composition
and nothing else.

\noindent\textbf{Result.} Three seeds per cell, ten epochs.

\begin{table}[!ht]
\centering\small
\begin{tabular}{@{}lccc@{}}
\toprule
 & native (paired), per seed & mean & probe($z$) \\
\midrule
$\rho = 0$ & $0.398$\quad$0.417$\quad$0.485$ & $0.434$ & $0.736$ \\
$\rho = 1$ & $0.471$\quad$0.485$\quad$0.529$ & $0.495$ & $0.708$ \\
\bottomrule
\end{tabular}
\caption{\textbf{Counterfactual support does not produce native reachability.} Native paired accuracy on What's~Up per seed and its mean, and a linear probe on the pooled vector $z$, with each anchor's exact left/right counterfactual in its batch ($\rho = 1$) and without ($\rho = 0$). The rule fixed before the runs counted $\ge 0.70$ as support and $\le 0.55$ as null.}
\label{tab:rho}
\end{table}

The decision rule was fixed before the runs: $\ge 0.70$ would have made this the paper's central
result and triggered a full $\rho$ sweep; $\le 0.55$ retires the hypothesis. The best single seed is
$0.529$.

\noindent\textbf{Why the null is interpretable.} A small corpus could fail to train at all, and a
model that learned nothing would also score at chance. The capability anchor rules that out:
probe($z$) is $0.736$ and $0.708$, inside the $0.63$--$0.75$ band spanned by the much larger
specificity arms. These models encode laterality as well as any encoder in the study, and still
cannot be read by the similarity metric.

\noindent\textbf{The rotation test, on the same checkpoints.} The counterfactual is a whole-image
mirror, so on its own it exercises the reflection axis rather than relation-confusability in general.
The $C_4$ orbits are its rotation-equivariant counterpart and cost no further training: they rotate
the displacement between two objects and leave the object pixels upright. On the six finished
checkpoints, the principal cosines between the image spatial subspace and the text relation subspace
are $\sigma_1 = 0.020$--$0.044$ and $\sigma_2 \le 0.014$, inside the range spanned by thirteen public
encoders that never saw a counterfactual ($\sigma_1 \le 0.083$, $\sigma_2 \le 0.024$), and native
4-way accuracy is $0.221$--$0.266$ against a chance of $0.25$, with $\rho{=}1$ no higher than
$\rho{=}0$. The \moment{} reads these same checkpoints at $0.318$--$0.368$: above chance, and far
below the $0.81$--$0.99$ the public encoders reach, which is what $65{,}536$ samples buys. So the
alignment does not rise on the axis the counterfactual was built for, and does not rise on the axis
it was not.

\noindent\textbf{Scope.} What is ruled out is that supplying exact same-content, opposite-relation
negatives is sufficient to make the lateral relation reachable, at this corpus size and this
schedule. It is the fifth lever in the series to leave native scoring at chance, after relational
caption frequency, phrasing diversity, phrasing held fixed, and caption truth.

\section{The pooled channel and the moment are different signals}
\label{app:bridge}

Per-encoder reader-direction cosines, their shuffled-label noise floors, and cross-fitted transfer.
The shuffled floor is the cosine between the pooled lateral reader and a reader of the same family
fitted on permuted labels; it is the right null because the cross-axis alternative (a reader fitted
on the \emph{same} lateral labels from vertical coordinates) is itself a lateral reader and cannot
say what $|\cos|$ counts as small at $n{=}206$ in $D \ge 512$.

\begin{table}[!ht]
\centering\scriptsize\setlength{\tabcolsep}{4pt}
\begin{tabular}{@{}lcccc@{}}
\toprule
Encoder & $\cos(w_{z^-}\!, w_{x})$ & shuf.\ p95 & mom$\to z^-$ & $z^-\!\to$mom \\
\midrule
clip-datacomp & $-0.037$ & $0.113$ & $0.466$ & $0.533$ \\
dfn-b & $+0.036$ & $0.117$ & $0.505$ & $0.612$ \\
clip-laion2b & $+0.047$ & $0.092$ & $0.544$ & $0.660$ \\
clip-openai & $-0.050$ & $0.141$ & $0.447$ & $0.471$ \\
metaclip-b & $-0.080$ & $0.103$ & $0.447$ & $0.243$ \\
convnext-clip & $-0.092$ & $0.170$ & $0.451$ & $0.353$ \\
siglip2-b & $-0.112$ & $0.173$ & $0.467$ & $0.448$ \\
rn50-clip & $-0.145$ & $0.150$ & $0.432$ & $0.277$ \\
siglip2-so400m & $\mathbf{+0.088}$ & $0.088$ & $0.557$ & $0.662$ \\
siglip2-l & $\mathbf{-0.278}$ & $0.131$ & $0.409$ & $0.185$ \\
\bottomrule
\end{tabular}
\caption{\textbf{The \moment{} and the pooled flip-odd probe read different directions.} Col.~2: cosine between the two lateral readers; col.~3: the 95th percentile of that cosine when the labels are permuted; cols.~4--5: accuracy of a reader fitted on one representation and scored on the other, cross-fitted. Bold: at or beyond its own shuffled floor.}
\label{tab:bridge}
\end{table}

Eight of ten sit inside their own floor. siglip2-l is significantly anti-aligned. siglip2-so400m
exceeds its floor on the positive side, at exactly its p95; it is weak, it is the only row anywhere
that supports a shared signal, and it is reported rather than rounded away.

The uniform evidence is the transfer failure: a lateral reader fitted on one representation and
scored on the other spans $0.185$--$0.662$, while each reads $0.77$--$0.99$ on its own. Three cells clear $0.60$ and none reaches what a reader achieves on the signal it was fitted for.

\section{What supervision does move}

\begin{table}[t]
\centering\small
\setlength{\tabcolsep}{4pt}
\begin{tabular}{@{}lccc@{}}
\toprule
Captions & seeds & native & $\cos(t_L,t_R)$ \\
\midrule
$p=0$ & 3 & 0.503 & 0.998 \\
$p=0.01$ & 3 & 0.494 & 0.794 \\
$p=0.05$ & 3 & 0.508 & 0.973 \\
$p=0.10$ & 3 & 0.510 & 0.956 \\
$p=0.25$ & 3 & 0.502 & 0.983 \\
$p=0.50$ & 3 & 0.481 & 0.984 \\
$p=1$ & 3 & 0.497 & 0.987 \\
\midrule
$p=1$, \emph{wrong rel.} & 3 & 0.510 & 0.948 \\
\bottomrule
\end{tabular}
\caption{\textbf{In these from-scratch runs, relational supervision does not produce native reachability, even at $p=1$.} Twenty-four CLIP-style encoders, three seeds per cell, trained from scratch on the same 373{,}528 images for the same $29.06$M samples seen, about $78$ passes over the corpus; $p$ changes captions only. Native lateral scoring stays at chance across the axis while the pooled vector stays decodable, and at $p=1$ the two text embeddings are still almost collinear. The wrong-relation control fixes relation-word frequency and randomises truth.}\label{tab:wp6}
\end{table}

\label{app:supmech}

Across the specificity axis, the three terms of Eq.~\ref*{eq:decomp} behave differently, and only one
of them responds. Encoding rises slightly: probe($z$) is $0.656$--$0.733$ at $p=0$ against $0.747$
at $p=1$, and the seed-resolved comparison resolves above zero ($+0.088$ over the wrong-relation
arm, $+0.046$ over $p=0$). Alignment does not: $|\cos\angle(\Delta\mu, \delta_T)| \le 0.19$ with the
sign varying between seeds, and its worst-arm seed spread ($\pm 0.121$) exceeds the range of its own
point estimates, so no ordering in that quantity is reportable. Native accuracy is likewise
unmoved: every pairwise difference is at most one item in $206$, the smallest value the measurement
can express, so those comparisons fall below resolution rather than merely short of significance.

Correct relational supervision therefore reaches the \emph{encoding} term and leaves the
\emph{alignment} term where it was. With two to three seeds per arm this is a direction, not an effect size.

\section{The phrasing and truth axes in full}
\label{app:axis2}

Fifteen arms, three seeds per cell, relational frequency held identical across the phrasing
conditions ($92{,}885$ captions) so that only phrasing differs.

\begin{table}[!ht]
\centering\footnotesize\setlength{\tabcolsep}{4pt}
\begin{tabular}{@{}lccc@{}}
\toprule
 & fixed & diverse & wrong rel. \\
\midrule
$p = 0.25$ & $0.4903$ & $0.4951$ & $0.4984$ \\
$p = 1.0$  & $0.4967$ & $0.5049$ & $0.5097$ \\
\midrule
$\cos(t_{\text{left}}, t_{\text{right}})$, $p{=}0.25$ & $0.962$ & $0.991$ & $0.979$ \\
\bottomrule
\end{tabular}
\caption{\textbf{Phrasing and truth do not move native reachability.} Native lateral accuracy, mean over three seeds, for fixed phrasing, diverse phrasing and wrong-relation captions at two relational frequencies $p$, and the text contrast $\cos(t_{\text{left}}, t_{\text{right}})$ at $p = 0.25$.}
\label{tab:axis2}
\end{table}

Condition B breaks one-token symmetry comprehensively: constructions that swap noun positions
(``on the left, $A$; on the right, $B$''), swap which object is named (``the leftmost object is the
$A$''), swap order (``scanning left to right, the $A$ then the $B$'') and change vocabulary (``$A$
is west of $B$''). The null is therefore not for want of a strong manipulation.

Condition C is the arm that makes the comparison readable. Its relation word is drawn independently
of the geometry, so it cannot teach the relation; if relational signal were being extracted at all,
C should fall below B. It does not: it is the tightest cell in the batch (seed sd $0.004$) and sits
nominally above the arm whose captions are all true. Arm C is what separates B's null from two effects cancelling.

The text-side direction is against the hypothesis rather than merely null: relation words are
\emph{more} collinear under diverse phrasing than under a single frame, and stay above $0.96$ in
every arm. Seed spreads overlap, so this is suggestive, not resolved.

\section{Relation planes, at the output and layer by layer}
\label{app:depth}

\noindent\textbf{Principal angles between relation planes.} A scalar cosine cannot express
anisotropy, so the natural generalisation is the \emph{principal angles} between a two-dimensional image
relation plane $\mathrm{orth}[w_x, w_y]$ fitted from What's~Up's minimal pairs and the
two-dimensional text relation plane
$\mathrm{orth}[t(\text{right}){-}t(\text{left}),\; t(\text{on}){-}t(\text{under})]$, whose cosines
are the singular values of $Q_I^{\!\top} Q_T$. It is only meaningful against an image plane that
carries the relation, which the $C_4$ plane does not, so we measure it on What's~Up, whose minimal
pairs read their own held-out instances ($0.767$--$0.990$ laterally, $1.000$ vertically). Both
principal cosines are near zero there ($\sigma_1 \le 0.239$, $\sigma_2 \le 0.077$, ten encoders), so
the near-orthogonality is a property of the whole plane, not of one measured direction.

\noindent\textbf{The text tower, layer by layer.} Everything above is measured at the output, which
leaves open that the tower builds the relation geometry at some depth and loses it by the last
layer, the shape \cite{negationlens} report for negation. Applying each encoder's \emph{own} pooling
head at every depth rules that out: across $147$ layers of the $8$ encoders \emph{that} a
pre-committed integrity check certifies, native $4$-way scoring stays at $0.226$--$0.311$ and
What's~Up lateral at $0.476$--$0.534$, every principal cosine lies inside its own permutation null,
and the relation contrast is estimated precisely enough at each depth to see one if it were there
.

The principal angles above measure the relation subspace at the output. If the tower built an aligned
geometry earlier and discarded it, every number in this paper would describe a late-layer artifact
rather than a property of the representation, and the repair would be attacking the wrong stage.
\cite{negationlens} report exactly that shape for negation, and \cite{negationwhere} localise
negation to particular depths in the same tower, so the possibility is live rather than hypothetical.

\noindent\textbf{The lens.} For a tower of depth $L$, layer $l$'s residual stream at the pooled
position is mapped into the joint space by the encoder's \emph{own} pooling head,
$t_l(c) = \mathrm{proj}(\mathrm{pool}(\mathrm{LN}_{\text{final}}(h^l(c))))$, which at $l = L$ is by
construction the deployed embedding. We do not fit anything: a tuned map would train parameters and
would measure our map instead of the model. Both towers here are covered, and they are not on the
same footing. The CLIP family is causal over $77$ tokens, pooled at the end-of-text position, with a
$(D,D)$ projection matrix; SigLIP2 is bidirectional over $64$ padded tokens, pooled at the last
position, with a projection that carries a bias. Depths are therefore compared as $l/L$.

\noindent\textbf{Why this is not the per-token move Sec.~\ref*{sec:method} rejects.} That section
reports that applying the final layer norm and projection \emph{per token} does not yield joint-space
vectors and is anti-correlated on CLIP. The concern there is cross-position: spatial tokens read
individually after the last attention block has already mixed them. Here the position is the one the
model itself pools from and only the depth changes, which is the standard intermediate-layer pooling head.

\noindent\textbf{Two anchors, both exact.} At $l=L$ the lens reproduces \texttt{encode\_text} to
$0.0$ on all ten encoders, and the resulting What's~Up 2AFC reproduces the published native lateral and on/under scores to $10^{-3}$. The second was not planned and is the stronger of the two,
since it exercises the whole scoring protocol rather than the embedding alone.

\noindent\textbf{Pre-registration.} Thresholds were committed before the panel ran. Firing required
either a layer scoring natively above chance on $\ge 6$ of $10$ encoders at agreeing relative depth,
or $\sigma_1 \ge 0.20$ together with $\ge 3\times$ its final-layer value, significance against a
max-over-layers permutation null, and split-half reliability $\ge 0.5$. Neither fired. The largest
peak-to-output ratio observed was $3.58\times$ (convnext-clip) but at $\sigma_1 = 0.113$, and the
largest $\sigma_1$ anywhere was $0.142$ (clip-openai, layer $7$ of $12$) against a null 95th
percentile of $0.164$.

\noindent\textbf{Controls.} The permutation null keeps each layer's caption geometry and destroys
only the relation labelling, so it absorbs the strong depth-varying anisotropy of the caption cloud;
the statistic is the maximum over layers, which handles depth without a multiplicity correction.
Split-half reliability estimates $\delta_T$ from disjoint halves of the $300$ orbits: it is
$\ge 0.988$ at every layer of every encoder, because the relation contrast is a near-constant vector
independent of which objects the caption names. That is what makes the null a measurement rather
than an absence of power. As a lens-integrity check we trace on/under, a relation some encoders
partially read at the output, through the same lens; on clip-openai it climbs smoothly to exactly its
published $0.607$ with no interior peak.

\noindent\textbf{Two encoders are not certified.} On siglip2-l and siglip2-so400m the on/under trace
peaks one or two layers \emph{before} the output ($0.675 \to 0.612$ and $0.631 \to 0.578$), so the
lens is not demonstrably clean there and the rule returns \textsc{inconclusive}. Their lateral traces
are flat like the rest, but we do not count them. A third encoder, clip-datacomp, returned
\textsc{inconclusive} under the rule as literally committed because the on/under test was applied
even though that encoder is at chance on on/under at the output, where the test has no content and
responds to sampling noise ($n=206$). Both tallies are given, $7$ and $8$ of $10$, because the correction favours our own claim.

\section{Symmetry transport, and why it fails}
\label{app:transport}

The paper diagnoses the lateral failure as misalignment. Taken alone, that invites an obvious
remedy: rotate the text relation plane onto the image one and score the original pooled vectors.
The remedy is attractive because it would need no patch grid, no re-indexing and no training,
and because it would validate the diagnosis using the very signal the diagnosis is about. It fails,
and the reason is the finding.

The thirteen encoders here are the ten of Tab.~\ref*{tab:main} plus CLIP ViT-B/32 LAION-2B and the
two TIPS-v2 models, added because a spatially-supervised objective is the case where a text-side repair
has the best prior chance of working.

\noindent\textbf{Construction.} From the $300$ $C_4$ orbits, $U_I = \operatorname{orth}[w_x, w_y]$
with $w$ either the frequency-1 channels themselves or their Fisher directions, and
$U_T = \operatorname{orth}[t_x, t_y]$ from the orbit captions. Both frames are built in the same
order and signed against their own generators, so column $j$ of each names the same relation axis and
the rotation between them is the identity in these coordinates: the correspondence comes from the
group action, not from a fit. The relation coefficient $a = U_T^{\!\top}(t - \mu)$ is centred on a
generic COCO caption mean, because the shared content of a raw caption also projects onto the
relation plane and that offset is identical for ``left'' and ``right''; it does not cancel in the
paired protocol, where the caption is held fixed. The transport is
\begin{equation}
  A = I - U_TU_T^{\!\top} + U_IU_T^{\!\top}.
  \label{eq:transport}
\end{equation}

\noindent\textbf{Propositions.} For $A$ as in Eq.~\ref{eq:transport}: if $v \perp
\operatorname{span}(U_T)$ then $Av = v$, so only the diagnosed subspace moves; if $v = U_Ta$ then
$Av = U_Ia$, so the relation plane lands exactly on the image plane; and
$A - I = (U_I - U_T)U_T^{\!\top}$ has rank at most $2$, which is minimal for a linear map
transporting a generic $2$-D subspace. All three hold numerically on the real bases to $1.8\times
10^{-8}$.

\noindent\textbf{Pre-registration.} The gate was What's~Up paired lateral $\ge 0.70$ against a
published baseline of $0.500$, judged on one variant named in advance, the $\alpha \to \infty$
centred channel score that the diagnosis predicts, rather than on the maximum over roughly thirty
variant--basis--$\alpha$ combinations. Nothing fired on any encoder. For completeness the other variants are reported too, and the
$\alpha$ sweep from $0.5$ to $25$ is flat.

\noindent\textbf{The premise, tested without text.} A transport built on one stimulus set and
applied to another needs the relation to occupy the same image-side direction in both. It does not.
Three directions were compared per encoder: fitted inside What's~Up on paired differences and
cross-validated by held-out object pair; built from What's~Up's own horizontal mirrors; and
estimated from the $C_4$ orbits.

\begin{table}[!ht]
\centering\footnotesize\setlength{\tabcolsep}{2.5pt}
\begin{tabular}{@{}lcc@{}}
\toprule
image-side lateral direction & reads pairs & $|\cos|$ with real-pair dir. \\
\midrule
fitted in What's~Up, cross-fitted & $0.660$--$0.961$ & $1$ \\
What's~Up's own mirror contrast   & $0.621$--$0.951$ & $0.369$--$0.752$ \\
$C_4$ orbits, frequency-1         & $0.456$--$0.583$ & $0.003$--$0.112$ \\
$C_4$ orbits, Fisher              & $0.359$--$0.660$ & $0.004$--$0.069$ \\
\bottomrule
\end{tabular}
\caption{\textbf{Three image-side lateral directions do not coincide.} Accuracy of each direction on What's~Up photo pairs, and its $|\cos|$ with the direction fitted on those pairs. Ranges over the ten encoders.}
\label{tab:transportdirs}
\end{table}

The middle row is the control that matters. A $C_4$ orbit member differs from its neighbour by a
rendered displacement of segmented cutouts, whereas a What's~Up pair is two different photographs,
so a small angle between those two could have meant only that the transformations differ. It does
not: a mirror contrast on What's~Up's own photographs is displacement-like, agrees with the
real-pair direction at $|\cos|$ up to $0.752$, and reads the pairs nearly as well, while the $C_4$
direction is near-orthogonal to \emph{that} as well ($|\cos| \le 0.106$). The split is by stimulus
set, not by transformation type.

\noindent\textbf{What this costs and what it buys.} It costs the cleanest possible validation of the
diagnosis. It buys the reason the remedy in Sec.~\ref*{sec:method} has the form it does: a
query-conditioned \moment{} recomputes the spatial geometry for each image and each query, so it never
needs a direction to be stable across stimulus sets. A stored direction does, and none is.

\section{Why not learn the pooling head?}
\label{app:attnpool}

\begin{table}[t]
\centering\footnotesize
\setlength{\tabcolsep}{3pt}
\begin{tabular}{@{}lccccc@{}}
\toprule
& \multicolumn{2}{c}{zero-parameter} & \multicolumn{3}{c}{learned attention pool} \\
\cmidrule(lr){2-3}\cmidrule(lr){4-6}
Encoder & rule & $+$probe & no coords & $+$coords & $+$coords \\
& (WU) & (WU) & (WU) & (in-dist.) & (WU) \\
\midrule
CLIP ViT-B/16 & 0.772 & \textbf{0.777} & 0.510 & 0.997 & 0.553 \\
CLIP LAION-2B & \textbf{0.806} & \textbf{0.806} & 0.519 & 0.999 & 0.466 \\
CLIP DataComp & 0.694 & \textbf{0.699} & 0.519 & 0.998 & 0.534 \\
MetaCLIP-B & \textbf{0.772} & 0.752 & 0.515 & 0.998 & 0.709 \\
DFN-B & \textbf{0.782} & 0.752 & 0.529 & 0.997 & 0.568 \\
SigLIP2-B & 0.859 & \textbf{0.966} & 0.500 & 0.999 & 0.500 \\
SigLIP2-L & \textbf{0.966} & \textbf{0.966} & 0.510 & 0.990 & 0.539 \\
SigLIP2-so400m & \textbf{0.990} & \textbf{0.990} & 0.524 & 1.000 & 0.573 \\
ConvNeXt-CLIP & \textbf{0.893} & 0.840 & 0.515 & 0.998 & 0.471 \\
RN50-CLIP & \textbf{0.743} & 0.670 & 0.505 & 0.999 & 0.471 \\
\bottomrule
\end{tabular}
\caption{\textbf{A learned pooling head solves the training distribution and does not transfer.} Query-conditioned readouts trained for 20 epochs on 47{,}000 lateral items from the composited corpus, then read on What's~Up (WU), which is photographs and which no arm has seen. Pool columns take the best of $k \in \{1,4,16\}$ queries, which favours the pool. Without coordinate channels it cannot express a \moment{} and sits at chance; with them it learns its own split and still does not transfer, except on MetaCLIP-B ($0.709$ $[0.660, 0.757]$, one arm of thirty).}\label{tab:attnpool}
\end{table}

The obvious alternative to a fixed rule is to learn one. We trained a query-conditioned attention
pool on $47{,}000$ lateral items from the controlled composited corpus for $20$ epochs and read it on
What's~Up, which it never sees. Tab.~\ref{tab:attnpool} reports both columns, and the in-distribution
one should be read first.

\noindent\textbf{The same head on the real-image corpus.} We also trained it on VSR and ViewSpatial
--- $1{,}742$ lateral items --- on six of the ten encoders, same architecture, same protocol, same
epochs.
There it does not fit: every arm is at chance on its own in-distribution held-out split
($0.440$--$0.545$ over thirty-six arms), with coordinate channels or without, so its
out-of-distribution numbers measure an untrained module and are not read as transfer. The
zero-parameter rule reads $0.762$--$0.990$ on those six. That is the reason the arms above are trained on composites.

\noindent\textbf{The comparison had to be rebuilt to be fair.} An earlier version trained a
$592$K-parameter pool inside each fold of a $206$-item benchmark and reported that it lost to the
\moment{}. A module that size on roughly $165$ training samples per fold is expected to overfit, so
that result showed a hand-designed inductive bias is more sample-efficient on a tiny set, not that
the \moment{} is better in general. Both sides here are query-conditioned, trained on the same pool
with the same group-disjoint split, and the \moment{}'s optional two-feature probe is fitted on that
same pool. Neither arm gains from where its parameters came from.

\noindent\textbf{Without coordinates the pool cannot express a \moment{}.} Attention pooling returns
$\sum_i w_i p_i$ with weights depending on token \emph{content}, so its output is a convex
combination of feature vectors and carries no position, exactly as a mean does. Reporting those arms
as a defeat for learned pooling would repeat the strawman above in a new costume, so we add a third
arm with coordinate channels appended per patch: $\sum_i w_i [\,p_i ; x_i ; y_i\,]$ contains
$\sum_i w_i x_i$, the query-conditioned centroid, so the architecture \emph{can} express what the
rule computes.

\noindent\textbf{It learns the distribution and does not transfer.} With coordinates the pool reaches
$0.980$--$1.000$ on its own held-out split, on all ten encoders, and $0.447$--$0.592$ on What's~Up on
twenty-nine of its thirty arms, while the zero-parameter rule reads $0.694$--$0.990$ on the same
items under the same protocol and is above the pool on every encoder. Capacity and optimisation are
not what stops it: it fits its own split almost perfectly, and across the sixty learned arms, which
span $0.8$M to $26.3$M parameters, there is no relation between size and out-of-distribution
accuracy --- SigLIP2-L's $26.3$M arm loses to DFN-B's $2.4$M one.

\noindent\textbf{The arm that transfers.} MetaCLIP-B with $k{=}1$ query and coordinates reads
$0.709$ $[0.660, 0.757]$ on What's~Up, well clear of chance and unlike anything else in the panel.
It is one cell of thirty and it is still below that encoder's zero-parameter rule
($0.772$ $[0.718, 0.825]$), the one encoder of ten where the two intervals overlap. It does not
change the comparison, but it does mean the pool's failure is not a law: a learned head can carry
something across this shift, and on nine encoders and twenty-nine arms of thirty it did not. What a learned pool acquires on one relational corpus is a way of
scoring that corpus; the rule carries no corpus with it, which is why it moves between stimulus sets
and the pool does not. This is the same split-by-stimulus-set that defeats symmetry transport
(Appendix~\ref{app:transport}).

\section{Nonlinear readers on the $C_4$ polarity}
\label{app:c4nonlinear}

A direction fitted on the orbits reads the polarity at chance, which shows it is not linearly
recoverable and not that it is gone. This repeats the measurement with readers that are not linear:
an MLP (two hidden widths), an RBF-SVM, and a cosine kNN. Groups are orbits throughout, folds are
\texttt{GroupKFold}, and every hyperparameter is chosen on an inner split of the training folds, so
no reported score has seen the value that picked it. Two designs are scored: the paper's own signed
differences $d = z_0 - z_2$ on the set $\{d, -d\}$, where a reader is through-origin and no class
prior exists, and the pooled vectors of orbit members $0$ and $2$ directly.

\noindent\textbf{Why this needs gates.} Six hundred rows in $512$--$1152$ dimensions is the regime in
which a flexible reader fails for want of data, and that failure looks identical to a null. Three
controls run on the same embeddings, the same rows and the same folds. \emph{Axis}: the same readers
on the task the linear probe already reads, at $0.911$--$0.992$. \emph{Planted-linear}: a label placed along
one random direction at amplitude $\varepsilon = \|u_x\|$ measured on that encoder's own orbits, so
the check certifies detection at the size the real polarity signal has rather than at a convenient
one; every reader returns $0.998$--$1.000$. \emph{Planted-XOR}: the exclusive-or of two planted coordinates
at the same amplitude, where neither coordinate alone predicts the label. A linear reader must sit at
chance and does, $0.447$--$0.522$; the three nonlinear readers return $0.978$--$1.000$. Both planted controls are run
twice, once on all $1{,}200$ pooled vectors and once on the very $600$ rows the polarity task is
scored on, so ``the reader ran out of data'' is excluded by an identical design with only the label
changed. Amplitudes are $\varepsilon = 0.102$--$0.173$ across the ten encoders.

\noindent\textbf{What the readers find.} Nothing the linear direction did not. Across ten encoders,
four readers and the two designs --- $80$ cells --- \textbf{one clears $0.60$}: RN50-CLIP under the
cosine kNN, at $0.613$ $[0.560, 0.670]$. RN50-CLIP is the encoder that already reads $0.587$ linearly and is named as
the exception in Sec.~\ref*{sec:c4}. The largest of the other $79$ cells is $0.587$, RN50-CLIP again. The rule
fixed before the data required two or more encoders above $0.60$ to change the account, so the
account stands, in the words the rule uses: no probe we trained, linear or nonlinear, recovers the
polarity from the pooled vector. That is a statement about a family of readers and a
finite sample, not about absence.

\noindent\textbf{The 4-way task.} The same readers score $0.354$--$0.522$ on the 4-way relation
against an axis-only ceiling of $0.5$, so the extra capacity does not buy polarity there either.

\section{Which stimulus families carry the polarity}
\label{app:frames}

A transfer claim needs a diagonal before it needs an off-diagonal. When a direction fitted on one
stimulus family fails to read another, either the relation occupies family-specific coordinates or
one of the families never carried a decodable signed direction, and only the second is true here.
This section measures, for every family, whether its own direction reads its own held-out instances,
and treats families that fail that test as ineligible for any transfer statement.

\noindent\textbf{Families.} Each supplies signed difference vectors whose correct sign is $+1$ by
construction, so a reader is a single direction through the origin and no convention bit exists.
\emph{wu-real}: the right-of photograph minus the left-of photograph of one What's~Up object pair.
\emph{wu-mirror}: a photograph minus its own horizontal mirror, signed by the relation.
\emph{wu-vert}: the on minus the under photograph of one pair. \emph{c4-disp}: $I_0 - I_2$, the same
two objects at the same two positions with the assignment swapped. \emph{c4-mirror}: $I_0$ minus its
own mirror, which places the objects exactly where $I_2$ does and differs from c4-disp only in
whether the object pixels are mirrored. \emph{c4-axis}: the control, $I_0 - I_1$ and $I_2 - I_1$,
which asks which axis the pair is arranged along rather than which polarity.

Directions are standardised logistic fits on $\{d, -d\}$, and the diagonal is cross-fitted with the
object pair or orbit held out, so no instance is scored by a direction fitted on itself.

\begin{table}[!ht]
\centering\footnotesize\setlength{\tabcolsep}{4pt}
\begin{tabular}{@{}lccc@{}}
\toprule
family & reads its own held-out instances & capable & chance \\
\midrule
wu-vert    & $1.000$          & $10/10$ & $0.500$ \\
wu-mirror  & $0.786$--$0.985$ & $10/10$ & $0.500$ \\
wu-real    & $0.767$--$0.990$ & $10/10$ & $0.500$ \\
c4-axis    & $0.998$--$1.000$ & $10/10$ & $0.500$ \\
c4-mirror  & $0.480$--$0.562$ & $0/10$ & $0.500$ \\
c4-disp    & $0.463$--$0.587$ & $0/10$ & $0.500$ \\
\bottomrule
\end{tabular}
\caption{\textbf{Which stimulus families carry a signed lateral direction in the pooled vector.} Each family's own direction, cross-fitted, on its held-out instances (range over the ten encoders), and on how many encoders it is capable, meaning it clears $0.60$.}
\label{tab:frames}
\end{table}

\noindent\textbf{Reading the table.} On real photographs the lateral polarity is in the pooled vector,
by two independent constructions that agree. On $C_4$ composites it is not, on any encoder, while
the arrangement axis is read essentially perfectly. Both $C_4$ families use the identical images and
differ only in how the contrast is formed, so this is not an artifact of one construction, and the
mirror variant holds content exactly fixed. The patch grid of those same images supports the full
4-way relation at $0.811$--$0.986$, so the polarity reaches the encoder and does not survive pooling.

\noindent\textbf{VSR is a third regime, measured not inferred.} VSR has no reversed-geometry
partner for a photograph, so the paired construction does not exist there; the mirror construction
does, and it is the same protocol. Signed by the relation each item asserts, it reads
$0.472$--$0.528$ and is capable on $0/10$. That is measured under the protocol of this section
rather than carried over from the pooled probe of Sec.~\ref*{sec:measure}, which reports the same
conclusion by a different route.

\noindent\textbf{Transfer, among the families that qualify.} The lateral direction moves between
What's~Up's mirror and real-pair constructions at $0.617$--$0.981$, so it is a property of the
family rather than of a single transformation. Directions taken from the $C_4$ families read the
What's~Up families at $0.398$--$0.631$, straddling chance (mean $0.498$), which carries no information given that those directions fail
their own diagonal.

\noindent\textbf{The background is not the explanation.} A composite on flat grey is unlike a
photograph, so the dissociation could be an artifact of the canvas. We rebuilt every orbit whose
source image was available locally ($207$ of $300$) with that orbit's own source photograph as a
blurred background, holding the foreground assets, scale, pair centre and displacement geometry
fixed, and compared against the \emph{same} $207$ orbits on flat grey rather than against the full
set. The polarity stays at chance either way ($0.420$--$0.628$ flat, $0.425$--$0.594$ photographic) while
the axis stays high ($0.995$--$1.000$ flat, $0.882$--$0.990$ photographic). The axis number does fall with a photographic background: some of the perfect axis reading
depended on the uniform canvas. The dissociation does not.

\noindent\textbf{One encoder is a partial exception.} RN50-CLIP, the only attention-pooled
convolutional model in the panel, reads the $C_4$ polarity at $0.587$ on the full set and $0.628$
$[0.560, 0.691]$ on the $207$-orbit subset, the single cell in $30$ that clears the capability bar.
Every other encoder is at chance on every $C_4$ polarity family. 

\noindent\textbf{The global-subspace analysis, family by family.} We fitted one relation subspace
on the union of all capable families except the one being scored, choosing the rank on half of that
family's groups and reporting the other half once, against a null that repeats the same
select-then-evaluate procedure on random subspaces so that both sides are maxima of the same shape.
The outcome splits by family type rather than by encoder. Axis families clear their null in $28$ of
$30$ cells, and signed-relation families in $1$ of $31$. The axis result is uninformative by
construction: \texttt{c4-axis}, \texttt{c4sub-axis} and \texttt{c4bg-axis} are built from the same
orbits, so leaving one family out does not leave its stimuli out. \textbf{Among the families that
actually carry a signed relation, held-out accuracy sits below the matched null in $30$ of $31$
cells, so we find no transferable global relation subspace.} 

\noindent\textbf{Two corrections this measurement forced.} The principal cosines of
Sec.~\ref*{sec:c4} were first computed against the $C_4$ image plane, which the table above shows
reads the relation at chance; a near-zero cosine against such a plane says nothing, and the
measurement now runs on the What's~Up plane instead. Separately, the $C_4$ vertical captions were
inverted relative to the placement arithmetic, since screen $y$ grows downward and $k{=}1$ places
$A$ above $B$ while the relation list said below. The images and the group action were correct, so
the fix is a relabelling; it leaves the principal cosines and the \moment{} unchanged, both being
polarity-insensitive or caption-free, and moves $C_4$ native 4-way from $0.233$ to $0.252$ on
\texttt{clip-openai}.

\section{Object size and position on What's~Up}
\label{app:size}

A lateral probe on What's~Up could in principle read a cue that travels with the relation rather
than the relation itself: relative object size, saliency, background asymmetry or viewpoint. We
tested the size version with a rule fixed before the run. OWLv2~\cite{owlv2} boxes both named
objects in every photograph ($1$ of the $103$ lateral pairs is excluded for an abstention, the same
pair the detector baseline abstains on). The pairs were then to be split by whether the caption's
subject is the larger object, and the flip-odd probe scored on each stratum, with the probe's own
out-of-fold predictions reproducing Tab.~\ref*{tab:main} on all ten encoders.

\noindent\textbf{The rule cannot fire.} No pair has a larger subject: the reference object is
always a chair, an armchair or a table, and the subject a smaller object placed beside it. By the
rule's own terms the result is \emph{underpowered}, because one stratum is empty.

\noindent\textbf{What the boxes show instead.} The reference object sits at the image centre, its
box centre at $0.48$--$0.52$ of the width, and moves at most $0.008$ of the width between a pair's
two photographs. Only the subject moves. Within a pair, then, size, background and viewpoint are
the same, and none of them can separate the two photographs. Neither can the larger object's side
of the frame, which as a predictor scores $0.500$. What does separate them is the subject's side of
the frame, which gives the label on $101$ of the $102$ pairs. The flip-odd probe on these pairs reads
$0.770$--$0.990$, its full-benchmark figures.

\noindent\textbf{What this means for the probe result.} On What's~Up the lateral polarity is
carried by the position of one object. A pooled vector can hold that without binding two objects to
their places, so the probe result shows that the pooled vector keeps where the moving object is,
which on these photographs decides the caption. It is not evidence of two-object binding. The $C_4$
orbits remove that cue, since both objects move between positions the pair shares, and there the
pooled vector loses the polarity (Sec.~\ref*{sec:c4}). This is also consistent with ConvNeXt-CLIP
reading What's~Up at $0.990$ from the flip-odd component while its average pool loses the polarity
on the orbits. The access failure is unaffected: the information that decides the caption is in the
pooled vector, and no caption we found reads it.

\noindent\textbf{One mislabelled pair.} In the published labels of the pillow/chair pair, the
pillow sits on the viewer's right in the photograph labelled \emph{left of}, and vice versa. We
keep the published labels throughout. That pair is the one exception among the $102$ above, and it
accounts for both of ConvNeXt-CLIP's errors.

\section{Standard caption retrieval with the parser in front}
\label{app:stdret}

\begin{table*}[t]
\centering\footnotesize
\setlength{\tabcolsep}{4pt}
\begin{tabular}{@{}lcccccccc@{}}
\toprule
& \multicolumn{4}{c}{Flickr30k test ($1{,}000$ images)} & \multicolumn{4}{c}{COCO val2017 ($5{,}000$ images)} \\
\cmidrule(lr){2-5}\cmidrule(lr){6-9}
& \multicolumn{2}{c}{all captions, R@1} & \multicolumn{2}{c}{parsed, R@1} & \multicolumn{2}{c}{all captions, R@1} & \multicolumn{2}{c}{parsed, R@1} \\
\cmidrule(lr){2-3}\cmidrule(lr){4-5}\cmidrule(lr){6-7}\cmidrule(lr){8-9}
Encoder & native & $+$moment & native & $+$moment & native & $+$moment & native & $+$moment \\
\midrule
CLIP ViT-B/16 & 0.592 & 0.587 & 0.638 & 0.514 & 0.302 & 0.295 & 0.305 & 0.205 \\
CLIP LAION-2B & 0.698 & 0.688 & 0.752 & 0.523 & 0.416 & 0.405 & 0.407 & 0.233 \\
CLIP DataComp & 0.676 & 0.665 & 0.725 & 0.486 & 0.398 & 0.388 & 0.404 & 0.250 \\
MetaCLIP-B & 0.707 & 0.702 & 0.748 & 0.619 & 0.414 & 0.406 & 0.409 & 0.295 \\
DFN-B & 0.691 & 0.677 & 0.766 & 0.445 & 0.427 & 0.414 & 0.431 & 0.235 \\
SigLIP2-B & 0.738 & 0.727 & 0.775 & 0.518 & 0.480 & 0.465 & 0.490 & 0.270 \\
SigLIP2-L & 0.771 & 0.759 & 0.830 & 0.555 & 0.503 & 0.490 & 0.506 & 0.298 \\
SigLIP2-so400m & 0.791 & 0.779 & 0.821 & 0.564 & 0.518 & 0.502 & 0.530 & 0.297 \\
ConvNeXt-CLIP & 0.721 & 0.711 & 0.766 & 0.518 & 0.420 & 0.410 & 0.432 & 0.269 \\
RN50-CLIP & 0.570 & 0.560 & 0.624 & 0.413 & 0.290 & 0.283 & 0.286 & 0.190 \\
\bottomrule
\end{tabular}
\caption{\textbf{On standard caption retrieval the spatial term costs R@1 on the captions the parser fires on.} Text-to-image R@1 with the rule-based parser in front of every caption and $\beta$ estimated out of domain on CC3M. The parser returns a triple for $4.4\%$ of Flickr and $6.5\%$ of COCO captions, almost all vertical; the rest score exactly as native, so \emph{all captions} differs from native only through the \emph{parsed} ones.}
\label{tab:stdret}
\end{table*}

Tab.~\ref*{tab:wp5} gives each query its triple and ranks candidates that already hold the named
objects. A deployed system sees ordinary captions, some of which happen to contain a relation word,
and ranks the whole pool; the guarantee of Sec.~\ref*{sec:method} covers only captions that name no
relation. We therefore put the rule-based parser of Appendix~\ref{app:freeform} in front of every
caption of two standard benchmarks: COCO val2017 ($5{,}000$ images, $25{,}000$ captions; not the
Karpathy $5$K test split) and Flickr30k test ($1{,}000$ images, $5{,}000$ captions), on all ten
encoders. $\beta$ is the CC3M estimate of Appendix~\ref{app:beta}, because the label-free
$\hat\beta$ was estimated from COCO val2017 captions and would otherwise have seen this pool. Native
R@1/5/10 reproduce our own earlier native runs within $0.0006$ on the eight encoders those cover.

\noindent\textbf{What the parser fires on.} $4.4\%$ of Flickr and $6.5\%$ of COCO captions, the
same captions for every encoder, almost all vertical: \emph{over} alone is $123$ of the $218$ Flickr
parses. Left/right fires on two captions per pool, and both COCO ones read a road's side as an object
(\emph{on the right side of a road}). In an audit of $100$ sampled parses the relation is genuine in most, but the head-noun
rule often returns a verb, a pronoun or a scene word as an object (\emph{sky above bridge} for two
planes flying over one), and some uses of \emph{over} are idiomatic (\emph{looking over}). This test
therefore speaks to the vertical axis, on natural captions.

\noindent\textbf{What it costs.} Unparsed captions score exactly as native, with a largest absolute
score difference of $0$ in float32 and no change of order, so \emph{all captions} in
Tab.~\ref{tab:stdret} moves only through the \emph{parsed} ones. There R@1 falls on every encoder,
and over all captions it drops by $0.005$--$0.015$, every bootstrap interval over captions
excluding zero. Parser error is not the cause: on the parses the audit reads as correct, R@1 falls
further than on those with a wrong object term. The spatial term has the right sign, and reversing
it is worse still. Across a whole pool, though, it is not selective: the gold image's spatial term
sits at the $61$st--$71$st percentile of the pool on average, and the term's spread over images,
$0.019$--$0.043$, is about the size of the native gap between the gold image and its nearest
competitor. Tab.~\ref*{tab:wp5} removes that competition by holding the named objects fixed, which
is why the same term ranks well there. With $\beta$ at a quarter of the estimate, parsed R@1 is at most
$0.073$ below native, by a median $0.019$; that is an observation on this test set, not a setting
selected for it.

\end{document}